# Action-grounded tissue affordance enables anticipatory auto-framing that lowers surgeon cognitive workload during laparoscopic surgery

Jiayu Gu[1,*], Yiwei Wang[2,5,*], Jie Zhang[2,*], Guojun Cao[1,*], Keshen Lyu[1], Song Zhou[2], Yimeng Chen[1], Haorui Wang[3], Qingmin Feng[4], Shenchao Shi[6], Hongkuan Shi[7], Qiuyu Yu[7], Qiang Xie[7], Huan Zhao[2], Wenbin Chen[2,5], Caihua Xiong[2,5], Chidan Wan[1], Jing Samantha Pan[3], Xiong Cai[1], Han Ding[2]

**Affiliations:**

1. Department of Hepatobiliary Surgery, Union Hospital, Tongji Medical College, Huazhong University of Science and Technology, Wuhan, China

2. School of Mechanical Science and Engineering, Huazhong University of Science and Technology, Wuhan, China

3. Department of Psychology, Sun Yat-sen University, Guangzhou, China

4. Institute of Biomedical Engineering, Union Hospital, Tongji Medical College, Huazhong University of Science and Technology, Wuhan, China

5. Institute of Medical Equipment Science and Engineering, Huazhong University of Science and Technology, Wuhan, China

6. Department of Hepatobiliary and Pancreatic Surgery, Renmin Hospital of Wuhan University, Wuhan, China

7. Wuhan United Imaging Surgical Co., Ltd. (United Imaging Surgical, UIS), Wuhan, China

Author e-mails: Hongkuan Shi, hongkuan.shi@ui-surgical.com; Qiuyu Yu, qiuyu.yu@ui-surgical.com; Qiang Xie, qiang.xie@ui-surgical.com

**Equal Contribution:**

* These authors contributed equally: Jiayu Gu, Yiwei Wang, Jie Zhang, Guojun Cao.

**Corresponding authors:**

Han Ding, Ph.D.

Address: No. 1037 Luoyu Road, Wuhan 430074, Hubei Province, China

Telephone/Fax number: 86-13886199756

E-mail: dinghan@hust.edu.cn

Xiong Cai, MD, Ph.D.

Address: No. 1277 Jiefang Ave, Wuhan 430022, Hubei Province, China

Telephone/Fax number: 86-15002748078

E-mail: caixiong@hust.edu.cn

Jing Samantha Pan, Ph.D.

Address: No. 132 Waihuan Donglu, Guangzhou 510006, Guangdong Province, China

Telephone/Fax number: 86-18922370280

E-mail: Panj27@mail.sysu.edu.cn

## Abstract

In laparoscopy, surgeon gaze tracks where the instruments will act; easing this demand through visual attention modeling requires dense labels of those interaction loci. These encode tacit knowledge: experts converge on consensus loci yet struggle to state the rules. Here we show that such labels can be recovered from completed actions in surgical videos, in which recorded instrument trajectories are converted into dense, continuous supervision. DiffeoAfford grounds tissue affordance by attaching instrument tips to the tissue and transporting them through deformation using diffeomorphism-constrained tracking, matching context-informed annotators' accuracy. Trained on these labels and never on gaze, a real-time model aligns with surgeon gaze more closely in space and time than does camera-assistant gaze. The framework also transfers across procedures: on hysterectomy videos, a separately trained predictor reaches 95.16% directional consistency with subsequent camera motion. In 12 paired cholecystectomies (24 procedures), the auto-framing application AffordView, which proactively centers predicted targets in view, lowered surgeon cognitive workload on converging subjective, physiological, and behavioral measures, including a reduced number of verbal instructions to the camera assistant. Deriving supervision from action rather than manual annotation offers a scalable route to anticipatory assistance.

## Keywords

Laparoscopic Surgery, Laparoscopic Self-regulation, Surgical Data Science, Visual Attention Modeling, Tissue Affordance

## Introduction

The American Medical Association (AMA) conceptualizes augmented intelligence as a framework that amplifies, rather than replaces, human cognitive capabilities[1, 2]. Advancing digital medicine from passive data analysis to proactive augmented intelligence requires artificial systems capable of overcoming sensory information overload[3]. Modeling visual attention and predicting regions of interest (ROI) are therefore foundational for bridging raw sensory data and high-level cognitive reasoning[4, 5, 6, 7]. Establishing this perceptual foundation facilitates the development of diverse augmented intelligence technologies, including attention-guided lesion detection, context-aware augmented reality, and human–robot collaboration[8, 9, 10, 11].

Laparoscopic surgery has emerged as a primary domain for such intelligent assistive applications[3, 12, 13], yet transitioning these technologies into the operating room reveals significant technical barriers[14, 15]. Although the degradation of laparoscopic video by adverse optical conditions was once considered the primary barrier to computational modeling, modern computer vision models can overcome such visual noise when trained on sufficient annotated data[14, 15, 16]. Thus, the principal obstacle in the surgical setting is not the appearance of the image, but the availability of labels.

While natural image datasets benefit from the scalable crowdsourcing of nameable objects with defined boundaries, manual annotation in the surgical domain faces profound, domain-specific barriers. In laparoscopy, the spatial ROI is a dynamic target dictated by functional surgical intent, identifiable only through specialized procedural knowledge. This judgment is intrinsically tacit: experienced surgeons reliably converge on consensus interaction loci yet cannot articulate the visual or geometric rules behind them. Compounded by continuous soft-tissue deformation and indistinct anatomical boundaries, this tacit reliance precludes lay annotators and explicitly coded heuristic models alike. Nor does reducing the number of required labels remove the barrier, as in one-shot affordance grounding of deformable objects from a single annotated exemplar per category[17], since such approaches presuppose a nameable part inventory labelable without specialist knowledge. Therefore, generating dense, per-frame annotations across long surgical videos depends entirely on a scarce pool of expert surgeons, rendering manual supervision prohibitively expensive and fundamentally unscalable[13, 14, 18].

Hence, we propose that the necessary supervision for computational visual attention models can be derived from a source other than manual annotation. Because completed procedures provide an implicit, retrospective record of spatial priority, the recorded instrument trajectories of the surgeon can retroactively ground the functional ROI. This hindsight-oriented approach motivates a conceptual shift from static anatomical landmark identification toward an action-grounded interpretation of surgical intent[19, 20].

To formalize this action-grounded interpretation of intent, we introduce tissue affordance prediction as a computational framework for intention-aware visual attention (**Fig. 1**). Rooted in the ecological perspective that attention guides action, the framework holds that effective perception prioritizes interaction loci that afford manipulation rather than exhaustive semantic labeling[21]. Empirical evidence demonstrates that expert surgeons employ consistent visuomotor strategies, fixating on shared affordance regions prior to executing movements[22, 23]. If expert visual attention is substantially shaped by actionability rather than instrument position alone, anticipating that attention plausibly requires predictive affordance detection rather than reactive tool tracking[22, 23].

We therefore formulated a tissue-affordance framework that requires no expert affordance annotation. Within this overarching architecture, we developed DiffeoAfford (Diffeomorphic Affordance Grounding), an affordance grounding pipeline that employs a hindsight strategy to retrospectively deduce surgical interaction loci (**Fig. 2**). This pipeline integrates global transformations and diffeomorphic deformations to attach historical instrument tips to the tissue and transport them through deformation across every frame, automatically generating dense, continuous soft labels of tissue affordance and bypassing the reliance on expensive, per-frame expert annotation. These soft labels define spatial target regions through accumulated actionability instead of rigid anatomical categories. To convert this retrospectively grounded data into proactive augmented intelligence, we subsequently trained a deep learning model to predict real-time affordance hotspots from the surgical scene. We validated grounding against expert consensus on public data and against intraoperative gaze in LCET, a proprietary eye-tracking dataset of laparoscopic cholecystectomies, then benchmarked prediction against an established reactive instrument-tracking baseline. LCET provides intraoperative gaze as an independent behavioral criterion rather than a restatement of the training signal, and its synchronized surgeon and camera-assistant recording allows anticipation to be measured against a human camera operator.

To translate visual attention modeling into clinical utility, DiffeoAfford underpins AffordView, a proactive spatial framing application that embodies applied augmented intelligence technology to mitigate surgeon cognitive workload (**Fig. 1**). Displaced surgical targets impose extraneous cognitive workload on the surgeon[24, 25]; to mitigate this burden, AffordView provides ergonomically optimized visualization inspired by participant auto-framing technologies utilized in consumer video conferencing[26]. In current practice, this alignment is maintained by the camera assistant, who must infer the surgeon's target from the same image while operating the scope, and whose misreadings surface only when the surgeon corrects them verbally. AffordView addresses that dependency by deriving the framing target from predicted tissue affordance rather than from a second person's reading of intent, proactively centering the surgical interaction locus and aligning the operative field with the optimized viewing position[20, 27]. The capacity of AffordView to mitigate cognitive workload was verified during paired surgical sessions using subjective workload questionnaires, continuous electroencephalography, and pupillometry metrics[28, 29].

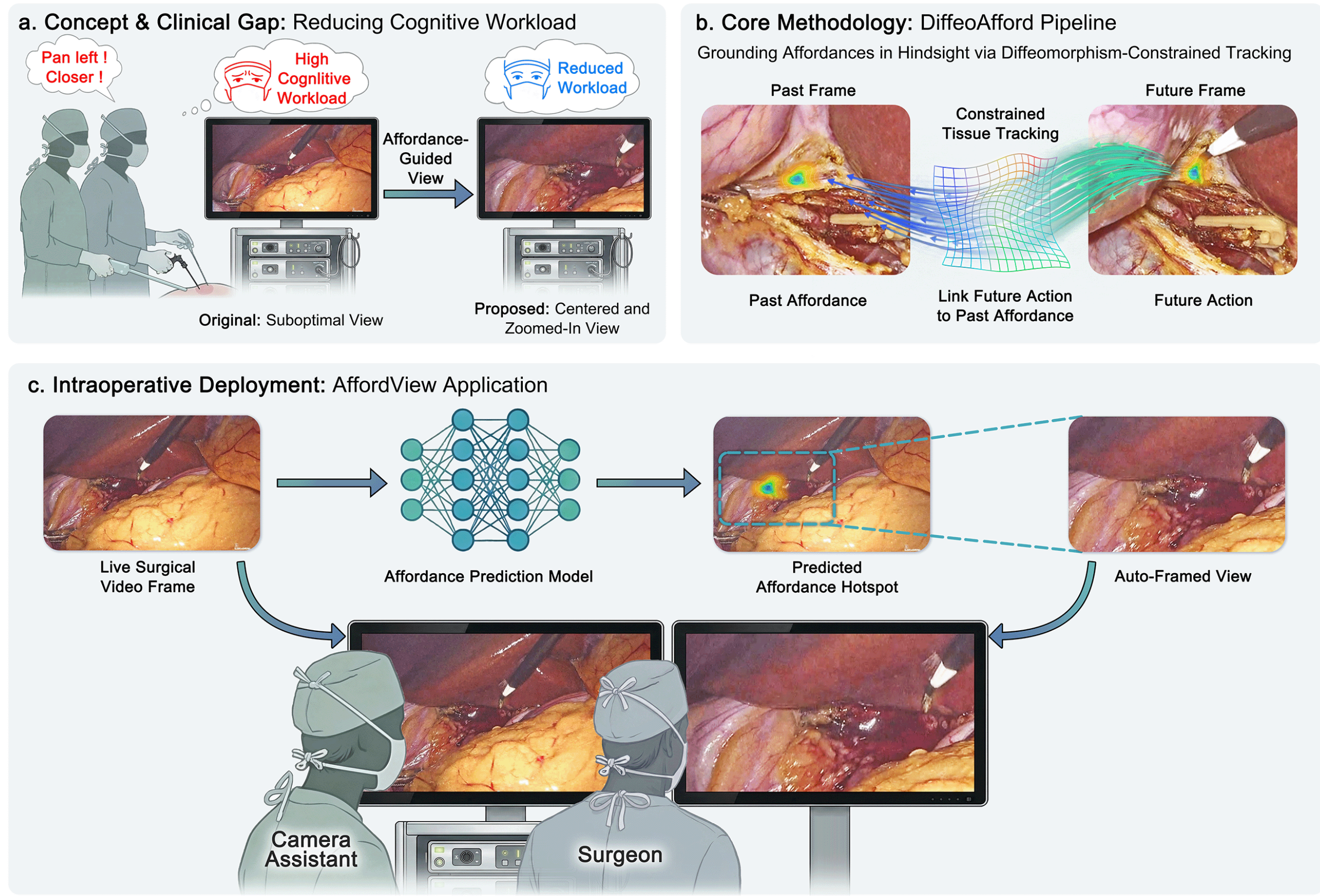

**Fig. 1 | Conceptual overview of the study. a**, **Concept & Clinical Gap:** Suboptimal field-of-view selection

elevates the surgeon's cognitive workload, which is mitigated through affordance-guided view adjustment. **b**, **Core Methodology:** DiffeoAfford, a pipeline that grounds affordances in hindsight via diffeomorphism-constrained tissue tracking to link future actions to past affordances. **c**, **Real-time Deployment:** AffordView, an application that predicts affordance hotspots to auto-frame the intraoperative view, providing an assistive view.

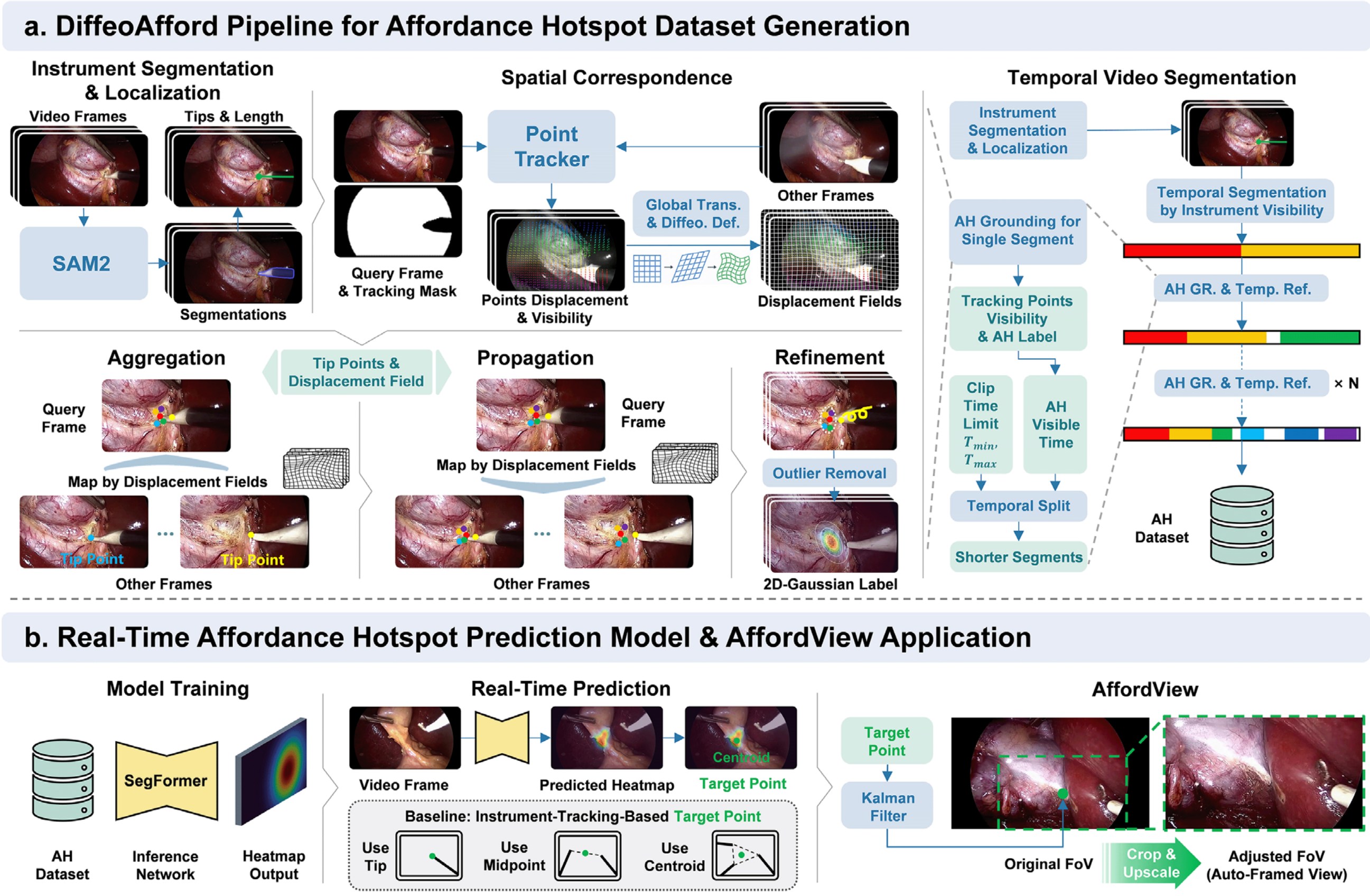

**Fig. 2 | Overview of the proposed affordance-guided view control framework.** The framework comprises DiffeoAfford pipeline for automated annotation (a), an affordance prediction model and the AffordView application for auto-framing (b). **a, DiffeoAfford pipeline for affordance hotspot dataset generation.** The SAM2 is utilized to segment instruments and derive instrument tips and lengths. To address soft-tissue deformation, DiffeoAfford performs diffeomorphism-constrained tissue tracking by combining global transformations with local deformation fields to establish dense pixel-wise correspondences across frames. Instrument tip points are aggregated onto a query frame, propagated to all other frames via displacement fields, filtered, and fitted to a 2D Gaussian distribution to generate AH heatmap labels. Video clips are then automatically segmented and refined based on instrument length and AH visibility using defined temporal thresholds ($T_{min}$, $T_{max}$). **b, Real-time affordance hotspot prediction model and the AffordView application.** A SegFormer model, trained on the generated AH dataset, predicts AH heatmaps in real-time. The centroid of the predicted heatmap serves as the target for the application, which dynamically crops and upscales the original video to maintain the view on the relevant surgical interaction zone, overcoming the limitations of baseline instrument tracking. AH, affordance hotspot; SAM2, Segment Anything Model 2; FoV, field of view; Global Trans. & Diffeo. Def., global transformation and diffeomorphic deformation; AH GR. & Temp. Ref., AH grounding and temporal refinement.

## Results

We first confirmed that surgeons' gaze exhibits a pronounced center bias on laparoscopic monitors, establishing the physiological rationale for center-aligned framing. Building on this finding, we conducted a series of studies organized in two parts (**Fig. 3**).

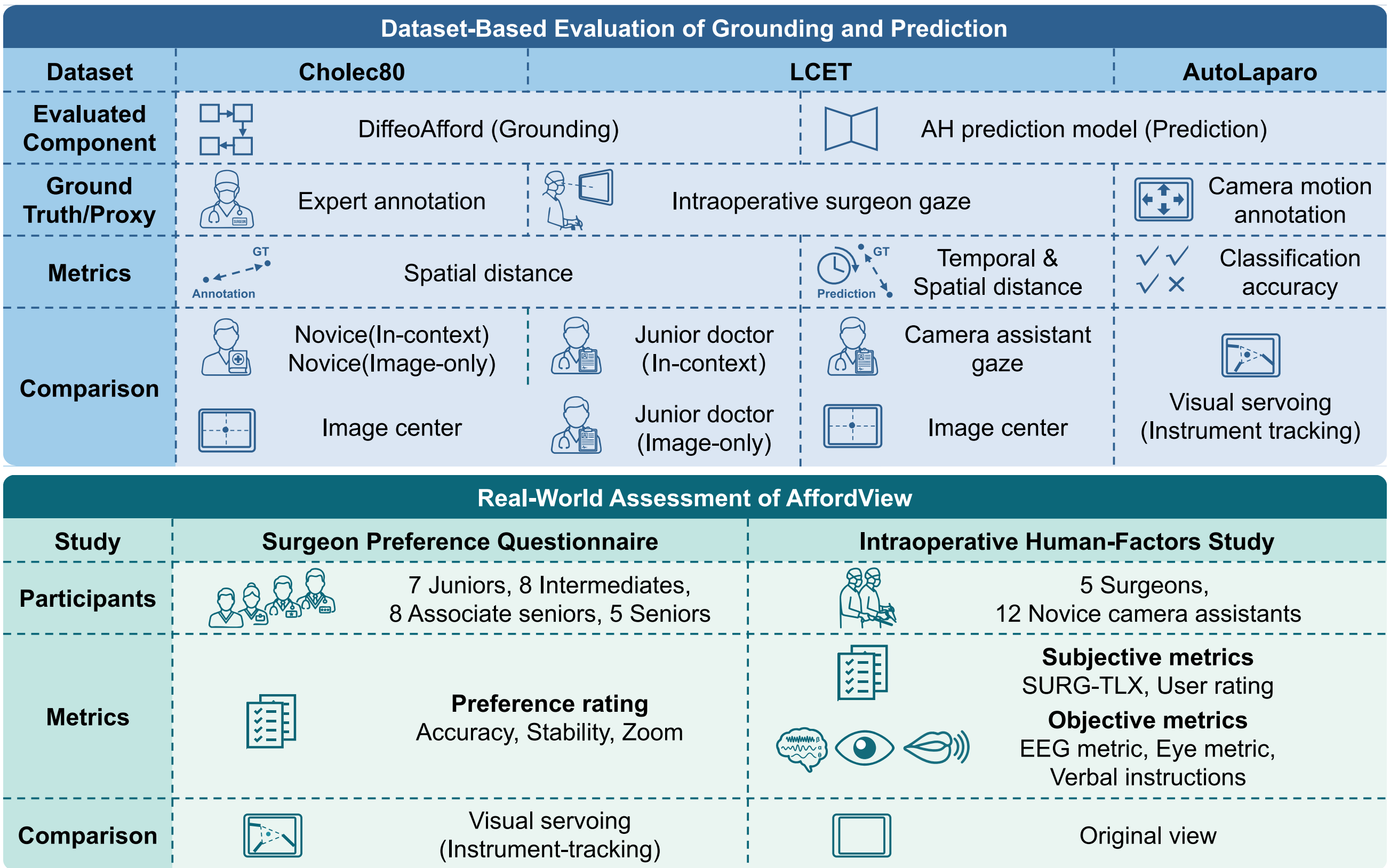

**Fig. 3 | Evaluation and assessment setup.** Overview of the study design for evaluation of the proposed method and assessment of the application. Dataset-based evaluations assessed DiffeoAfford grounding against expert consensus on Cholec80 and against intraoperative surgeon gaze on LCET, and AH prediction against synchronized surgeon and camera-assistant gaze on LCET and against subsequent camera motion on AutoLaparo. Application-level assessment comprised a multicenter surgeon-preference survey and an intraoperative human-factors study of cognitive workload.

### Center bias in laparoscopic surgery

To investigate whether the visual center bias reported in a previous study extends to the laparoscopic environment, we analyzed gaze data from our laparoscopic cholecystectomy eye-tracking (LCET) dataset ($n = 26$ procedures). Surgeons' gaze on laparoscopic monitors demonstrated a distinct center bias, as quantified by the Mahalanobis distance ($D_M = 0.05$) between the mean fixation point and the geometric center of the screen (**Supplementary Fig. 1**).

### Evaluation of DiffeoAfford on Cholec80 dataset

To test whether automatically generated affordance hotspots (AHs) align with human expertise, we evaluated DiffeoAfford on the public Cholec80 dataset. DiffeoAfford generated 141 clips from the Calot triangle dissection phase of Cholec80 videos 01–10; we sampled one eligible frame per clip and recruited three annotator groups: senior surgeons, junior doctors, and medical students (novices). Senior surgeon annotations served as the expert ground truth.

**Inter-annotator consistency:** We measured within-group consistency among human annotators using intraclass correlation coefficients (ICC (3, k)) (**Fig. 4a**). The consistency increased after annotators watched the video clips (hereafter referred to as the '**In-context**' condition), indicating that procedural context provides information beneficial for target recognition. After viewing the clips, novice consistency rose for both coordinates and approached the senior surgeons' consensus (**Fig. 4a**).

**Annotation accuracy:** Using senior surgeons' annotations as ground truth, we measured annotation error via Euclidean distance (**Fig. 4b**). To assess statistical differences, we employed linear mixed models (LMMs) with the Novice (In-context) group as the reference (**Fig. 4c**). The image center baseline exhibited an estimated error difference of 45.52 pixels (95% CI: 36.85–55.08, $p < 0.001$). Novices who viewed only static images showed a significant deviation from the reference (difference = 35.27 pixels, 95% CI: 29.74–41.20, $p < 0.001$), but their performance improved markedly after viewing clips. Notably, DiffeoAfford reproduced this context-aware performance. Its annotation error did not differ significantly from that of the Novice (In-context) group in the LMM (difference = −1.78 pixels, 95% CI: −5.76–2.68, $p = 0.418$). A bootstrap equivalence analysis confirmed equivalence with the Novice (In-context) group (**Supplementary Fig. 2c**). These results indicate that knowledge about the target is embedded in the procedural context, and DiffeoAfford-grounded AHs from this context successfully capture this expert knowledge.

**Improved affordance-grounding accuracy over global-transform baselines on Cholec80**

Previous affordance-grounding methods for natural-scene videos commonly propagate interaction points across frames using global transformations[30, 31]. Following this strategy, we constructed similarity RANSAC and homography RANSAC baselines, keeping all other pipeline components identical to DiffeoAfford (Methods). We compared the three methods on the same 141 Cholec80 images, DiffeoAfford achieved a lower median localization error (32.38 pixels) than both similarity RANSAC (39.40 pixels) and homography RANSAC (35.07 pixels) (**Supplementary Fig. 3**). Both differences were statistically significant (two-sided paired Wilcoxon signed-rank tests; similarity RANSAC, $p < 0.001$;

homography RANSAC, $p = 0.003$).

**Evaluation of DiffeoAfford against intraoperative gaze on LCET dataset**

To evaluate how well the generated AHs align with intraoperative intent, we used intraoperative surgeons' gaze points as proxy ground truth[32, 33, 34]. We calculated the Euclidean distance between these actual gaze points and the annotations across three distinct temporal windows (±0.5 s, ±1.0 s, and ±2.0 s) centered around the target frame (**Fig. 4d**). Statistical differences were assessed using LMMs following a Box-Cox transformation to satisfy residual normality. Using the Junior (In-context) group as the reference, we found that annotations of DiffeoAfford reached comparable accuracy across all three windows. Differences were not significant in any window, and a bootstrap equivalence analysis confirmed equivalence (**Supplementary Fig. 4g**). This indicates that DiffeoAfford-grounded AHs align with where surgeons intend to interact and can serve as a proxy for visual attention. The observation aligns with evidence that while free viewing is driven by general informativeness, attention in specific tasks becomes directed toward object parts that afford interaction[35, 36].

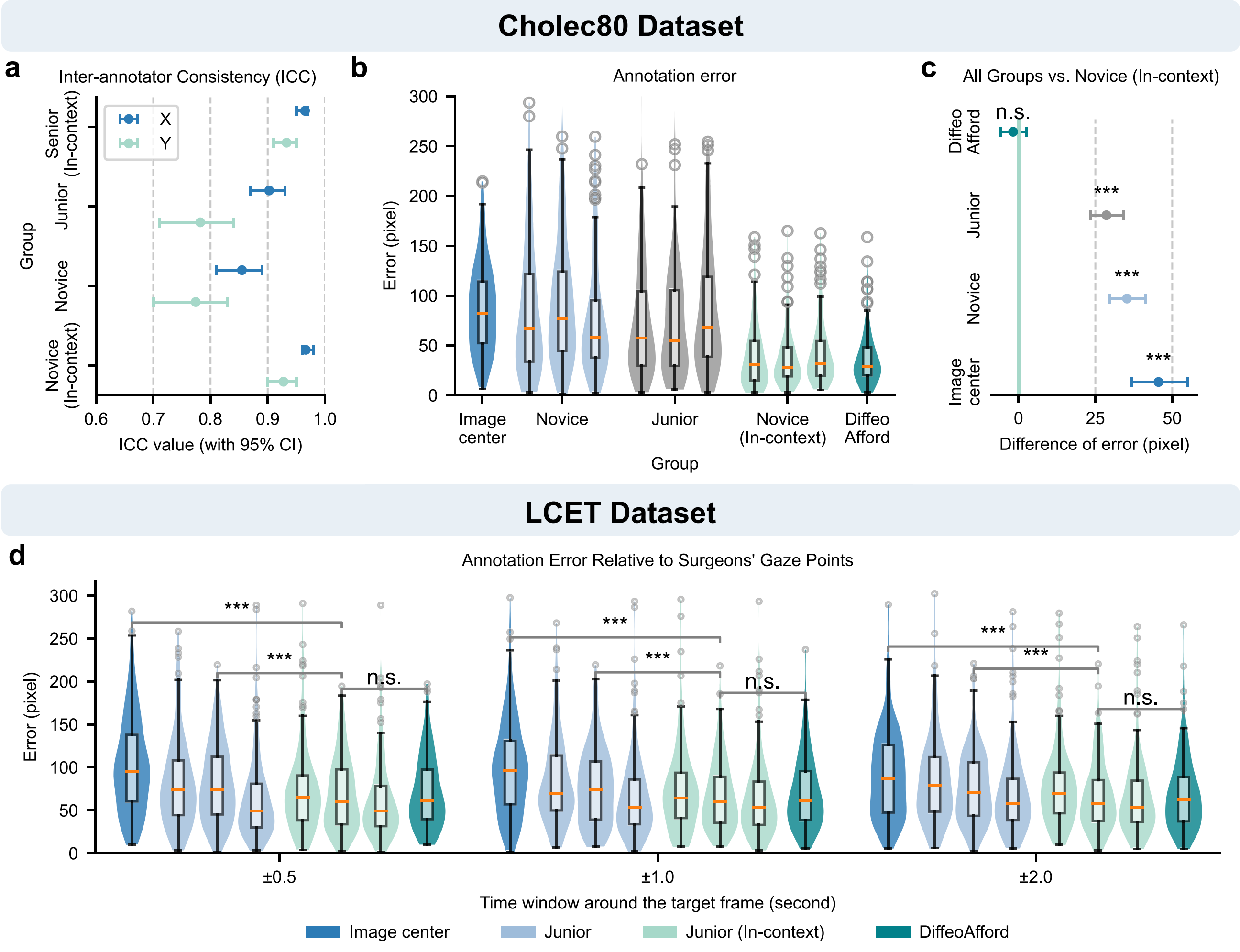

**Fig. 4 | Evaluation of affordance grounding across surgical datasets. a–c**, DiffeoAfford on the Cholec80 dataset ($n = 141$ images). **a**, Inter-annotator consistency for X and Y coordinates across participant groups, quantified by the ICC. Error bars denote 95% CI. **b**, Annotation error (in pixels) relative to the expert ground truth for the human annotator groups, the image center baseline, and DiffeoAfford. **c**, Forest plot of the estimated inter-group differences in annotation error relative to the Novice (In-context) reference group, presented as estimated differences ± 95% CI from linear mixed models (LMMs) fitted after a Box-Cox transformation to satisfy the assumption of residual normality. **d**, Annotation error relative to surgeons' intraoperative gaze points on the LCET dataset ($n = 164$ images). The violin plots display distances between annotations and the surgeon gaze points, which serve as proxy ground truth, across temporal windows (±0.5 s, ±1.0 s, and ±2.0 s) centered around the target frame; statistical differences were calculated using LMMs following a Box-Cox transformation. Inter-group statistical comparisons and LMM residual diagnostics are provided in **Supplementary Fig. 2a, b** (Cholec80) and **Supplementary Fig. 4a–f** (LCET); median-based bootstrap equivalence analyses of DiffeoAfford versus Novice (In-context) and versus Junior (In-context) are shown in **Supplementary Fig. 2c** and **Supplementary Fig. 4g**, respectively. For all violin plots (**b**, **d**), the internal box limits indicate the interquartile range (IQR), orange lines denote the median, and whiskers extend to a maximum of 1.5 × IQR, with circles representing individual outliers. $p$ values were adjusted for multiple comparisons using the Benjamini-Hochberg procedure; ***, $p < 0.001$; n.s., not significant. ICC, intraclass correlation coefficient; CI, confidence interval; LCET, laparoscopic cholecystectomy eye-tracking; LMM, linear mixed model.

## Evaluation of AH prediction against intraoperative gaze on the LCET dataset

**Temporal alignment:** The AH prediction model, trained on Cholec80 with DiffeoAfford-generated annotations, was applied to the 13 LCET procedures carrying synchronized surgeon and camera-assistant eye tracking, and the predicted AH centroids were compared with surgeon gaze using diagonal cross-recurrence profiles (DCRPs) during both Calot triangle dissection and gallbladder dissection, with camera-assistant gaze serving as a human reference sequence (**Fig. 5a**). The group-mean DCRP for AH prediction peaked on the candidate-leading side, whereas camera-assistant gaze peaked on the candidate-lagging side. At the procedure level, the median centroid lags for AH prediction were −0.033 s and 0.013 s in the two phases, compared with 0.185 s and 0.204 s for camera-assistant gaze ($p = 0.014 \; and \; 0.014$). Thus, the centroid lag of AH prediction relative to surgeon gaze was significantly shorter than that of camera-assistant gaze in both phases. No consistent temporal precedence was observed between AH prediction and surgeon gaze.

**Spatial alignment:** We further compared the spatial distances from surgeon gaze to AH prediction, camera-assistant gaze, and the image center at simultaneous valid timestamps, summarized as a case-level median distance for each phase and comparison. The median case-level advantage of AH prediction over camera-assistant gaze was 55.5 pixels during Calot triangle dissection and 28.4 pixels during gallbladder dissection ($p = 0.0093\ and\ 0.0479$); relative to the image center, the median advantages were 60.8 and 99.1 pixels ($p = 0.0015\ and\ 0.0061$) (**Fig. 5b**). AH prediction was thus significantly closer to surgeon gaze than either camera-assistant gaze or the image center in both surgical phases.

**Evaluation of AH prediction for camera motion forecasting on AutoLaparo dataset**

To test transfer to another procedure, we used AutoLaparo, a laparoscopic hysterectomy dataset. We trained the real-time prediction model using DiffeoAfford-grounded AHs from Task 1 videos, and subsequently evaluated its performance on Task 2 ($n = 124$ camera motion events across four directions: up, down, left, and right), which is specifically designed for evaluating camera motion prediction. During this evaluation, we predicted AH locations prior to camera motion events and compared the predictive performance against an instrument-tracking baseline (detailed in the Supplementary Information).

**Prediction accuracy:** Consistent with previous work[37], we analyzed the consistency between the AH location and camera motion within the image coordinate system across four directions: up, down, left, and right. The AH locations predicted immediately before camera motion exhibited strong consistency with the subsequent camera movement, with directional consistency of 95.16% (**Fig. 5c**). This shows that the predicted AH effectively anticipates the target driving the view adjustment.

**Superior foresight:** We compared the model with the instrument-tracking target prediction serving as the baseline. As shown in **Fig. 5d**, the instrument-tracking method achieved similar accuracy near the motion event, but its accuracy degraded rapidly as the temporal gap increased. Specific analysis revealed that instrument tips are often already near the target at the moment preceding camera adjustments, accounting for the short-term advantage; however, this assumption no longer holds at earlier time points. In contrast, the AH prediction maintained stable accuracy over longer horizons, offering superior foresight and robustness. Representative examples illustrating the differences between the two approaches are shown in **Fig. 5e**.

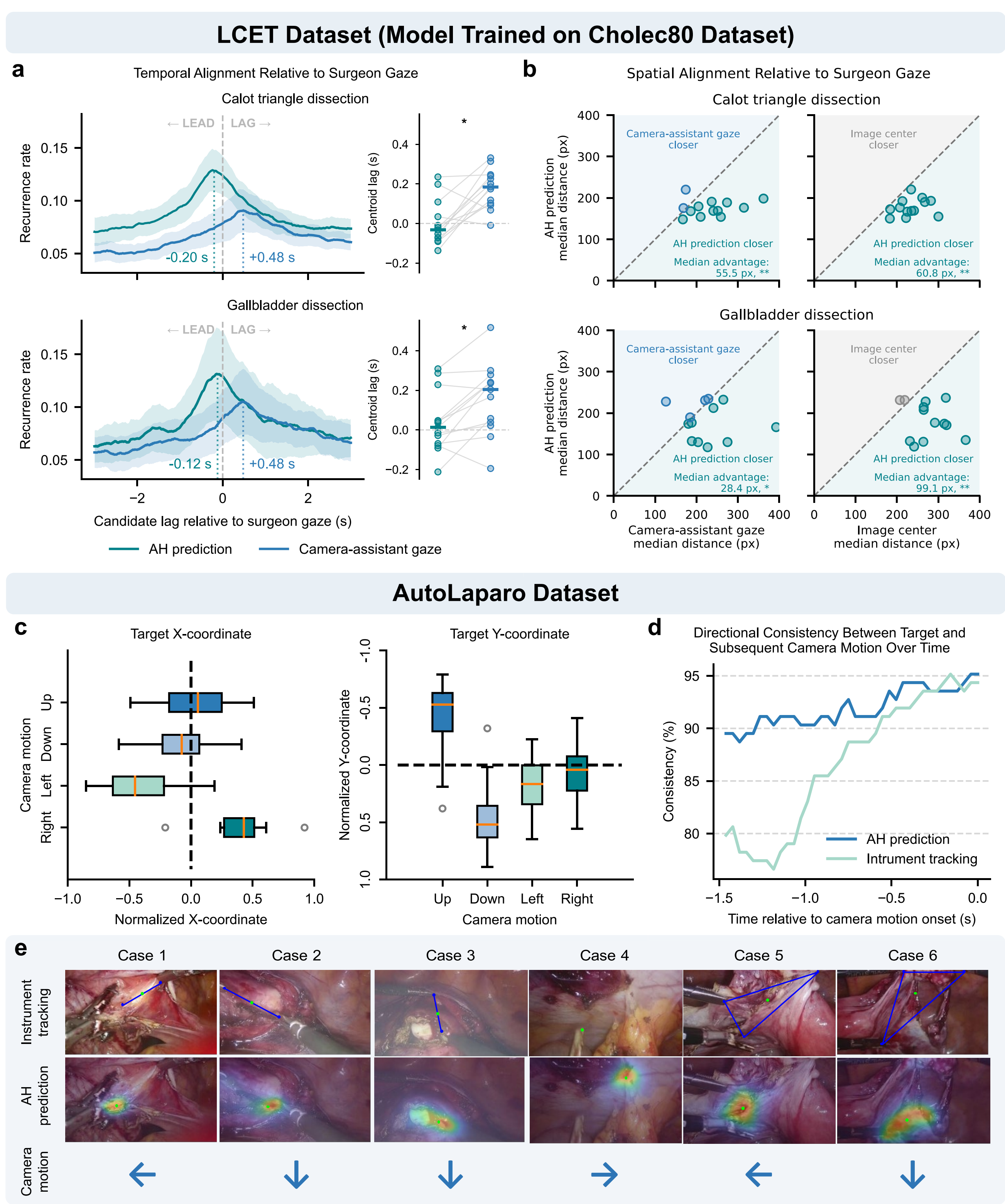

**Fig. 5 | Evaluation of AH prediction across surgical datasets. a, b**, Evaluation of the Cholec80-trained AH prediction model against intraoperative surgeon gaze in 13 LCET procedures with synchronized surgeon and camera-assistant eye tracking, during Calot triangle dissection and gallbladder dissection. **a**, Temporal alignment. Left, group-mean diagonal cross-recurrence profiles (DCRPs) obtained by comparing surgeon gaze at time $t$ with each candidate sequence at time $t + \tau$; negative lags indicate that the candidate sequence leads surgeon gaze, whereas positive lags indicate that it lags surgeon gaze. Solid lines denote the mean

recurrence rate across procedures, shaded bands denote mean ± 1.96 SEM, and colored dotted vertical lines mark the peaks of the group-mean DCRPs, which lay at −0.20 s during Calot triangle dissection and −0.12 s during gallbladder dissection for AH prediction and at 0.48 s in both phases for camera-assistant gaze. Right, procedure-level centroid lags, with points from the same procedure connected and horizontal bars indicating the medians; negative values indicate that the candidate tends to lead surgeon gaze, positive values indicate that it tends to lag surgeon gaze, and values near zero indicate temporal coupling without a consistent lead or lag. **b**, Spatial alignment. Each point represents one procedure, plotting the median distance from AH prediction to surgeon gaze (vertical axis) against the corresponding median distance from camera-assistant gaze or the image center to surgeon gaze (horizontal axis). The diagonal line denotes equal distances; points below the line indicate that AH prediction was closer to surgeon gaze. Median advantage denotes the median across procedures of the comparator's case-level median distance minus the corresponding AH-prediction median distance. For **a** and **b**, exact two-sided paired Wilcoxon signed-rank tests were applied to procedure-level values. **c–e**, Evaluation of AH prediction against subsequent camera motion in the AutoLaparo dataset ($n = 124$ clips with camera motion). **c**, Distribution of the normalized predicted AH X and Y coordinates at the onset of camera motion, categorized by the subsequent camera movement direction (up, down, left, right); the internal box limits indicate the interquartile range (IQR), orange lines denote the median, and whiskers extend to a maximum of 1.5 × IQR, with circles representing individual outliers. **d**, Directional consistency between the target (using the AH prediction model versus the baseline instrument-tracking method) and the subsequent camera motion over time, relative to the camera motion onset at the 5-second mark. **e**, Representative cases comparing instrument-tracking targets, AH prediction heatmaps, and subsequent camera-motion directions. Heatmap intensities were normalized independently to the maximum value in each panel for visualization. *, $p < 0.05$; **, $p < 0.01$; LCET, laparoscopic cholecystectomy eye-tracking; AH, affordance hotspot; DCRP, diagonal cross-recurrence profile; IQR, interquartile range; SEM, standard error of the mean.

### Surgeon preference assessment of AffordView application

Following the validation of the model's predictive accuracy, we integrated it into our AffordView application to assess its acceptability. We conducted a multicenter questionnaire survey in which participants compared paired Cholec80 clips: the original view versus AffordView, and AffordView versus the instrument-tracking baseline (Methods). Out of 32 distributed questionnaires, 28 valid responses were collected (an 87.5% response rate) from 21 tertiary-care hospitals. Respondent grade distribution is shown in **Fig. 6a**. As shown in **Fig. 6b**, the proposed method was significantly preferred over the original unadjusted view for FoV accuracy, favored in 77.1% (216/280) of the paired comparisons (Odds Ratio [OR] = 3.38, $p < 0.001$, evaluated via exact binomial test). However, no clear advantages were observed in FoV stability (48.6% vs 51.4%, OR = 0.94, $p = 0.676$) or zoom rationality (53.6% vs 46.4%, OR = 1.15, $p = 0.256$). Compared with the instrument-tracking baseline, the proposed method was preferred on all three dimensions: accuracy (75.4%), stability (75.7%), and zoom rationality (61.8%) (**Fig. 6c**). The lack of advantage in zooming prompted the integration of an instrument-driven zoom module for the subsequent intraoperative evaluation.

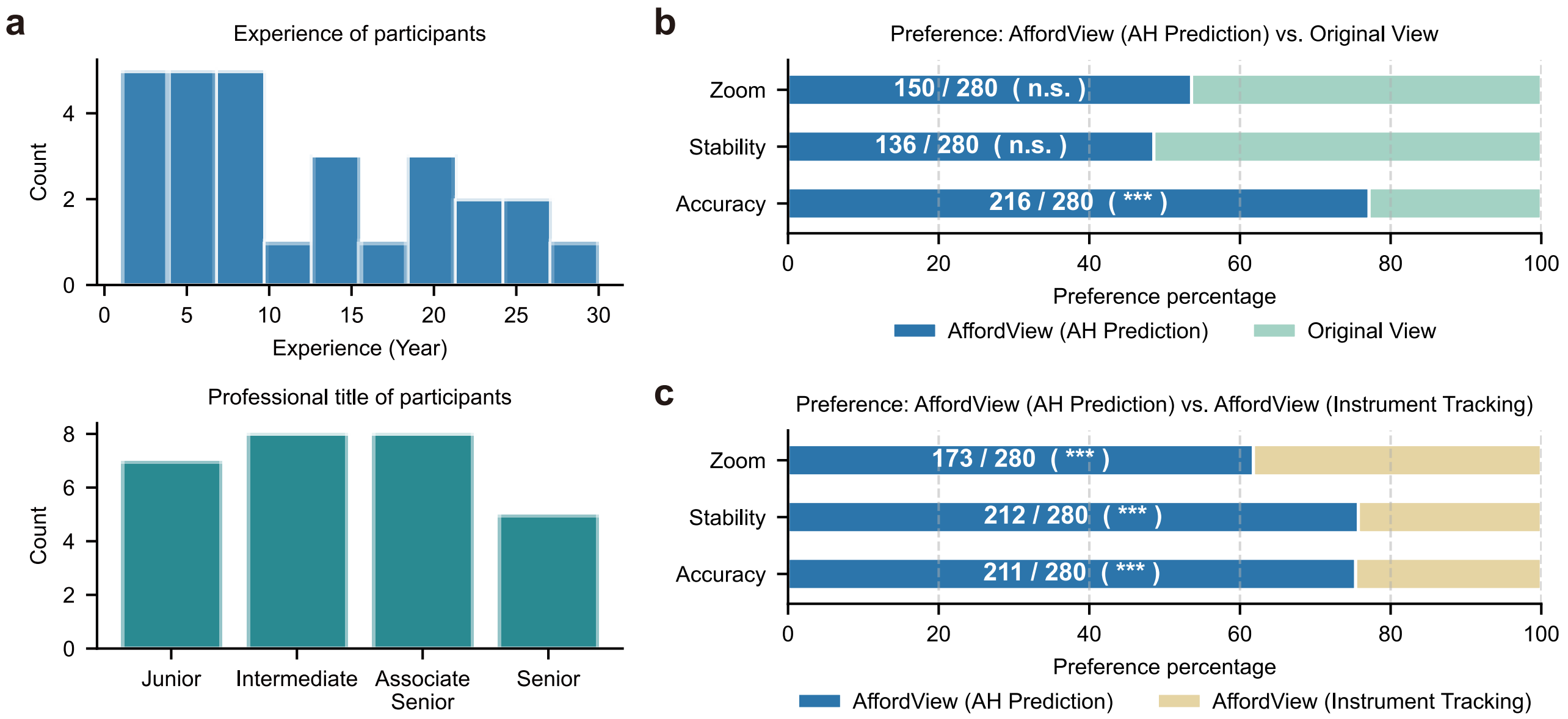

**Fig. 6 | Subjective assessment of the FoV adjustment. a**, Characteristics of the survey participants obtained from a multicenter questionnaire survey ($n = 28$ valid responses from 21 tertiary-care hospitals), illustrating the distribution of surgical experience in years (top) and professional titles (bottom). **b**, **c**, Surgeons' preference percentages comparing the FoV adjusted by the AH-prediction-driven AffordView against the original unadjusted view (**b**) and the baseline instrument-tracking-driven AffordView (**c**). Preferences were evaluated across three dimensions: zoom rationality, FoV stability, and FoV accuracy. The exact fraction of votes favoring the AH prediction method over the total comparisons is displayed within the bars. Statistical significance was evaluated via exact binomial tests; odds ratios in c were 3.06 (FoV accuracy), 3.12 (FoV stability) and 1.62 (zoom rationality). ***, $p < 0.001$; n.s., not significant.

## Intraoperative human-factors study of the AffordView application

To comprehensively assess the application's overall ergonomic impact in clinical practice, the AffordView application was deployed during real-world laparoscopic procedures (**Fig. 7**). Twenty-four laparoscopic cholecystectomies were performed by 5 operating surgeons and 12 novice camera assistants, comprising 12 matched pairs (each pair consisting of a control session and an experimental session). The median analyzed operative duration was 24.7 min (IQR, 16.6–46.2 min). The model was trained on the Calot triangle dissection (Calot) and gallbladder dissection (Diss) phases, the primary intervention targets. Given the relatively small sample size inherent to such highly controlled, intraoperative dual-monitor evaluations ($n = 12$ pairs), exact permutation-based paired tests were utilized as the primary inference method to ensure robust, distribution-free hypothesis testing. Linear mixed models (LMMs) were additionally employed to further control for potential confounding effects. All 12 pairs yielded complete data for analysis, with no pairs excluded.

**Objective metrics:** We assessed the surgeons' CWL and operational fluency through electroencephalography (EEG), eye-tracking, and verbal instructions. The results demonstrated surgical phase specificity, aligning with the targeted intervention phases.

For the EEG analysis, the theta/alpha ratio (TAR) (**Fig. 7a**), a physiological marker of CWL[38, 39, 40], decreased in the experimental group during both phases (Calot: Mean difference = −0.40, Cohen's $dz = -0.64$, permutation $p = 0.023$; Diss: Mean difference = −0.42, Cohen's $dz = -0.86$, $p = 0.012$). Relative to controls, occipital alpha power in the experimental group declined during the later surgical phases, with a significant reduction during the gallbladder dissection phase (Mean difference = −0.90, Cohen's $dz = -0.67$, $p = 0.043$) (**Fig. 7c**). This pattern is compatible with increased visual engagement[41, 42]. The EEG engagement index (EI), which reflects general alertness and focus[43, 44], showed no consistent phase-specific difference (**Fig. 7b**).

Parallel improvements were observed in oculomotor parameters: the index of pupillary activity (IPA)[45], a physiological marker of CWL, decreased during the two intervention phases (**Fig. 7f**) (Calot: Mean difference = −0.037, Cohen's $dz = -0.79$, $p = 0.021$; Diss: Mean difference = −0.040, Cohen's $dz = -1.43$, $p = 0.002$). Within the experimental sessions, the proportion of fixation time directed to the AffordView monitor increased during the two target phases, indicating greater use of the adjusted view during active intervention (**Fig. 7e**). Notably, while stationary gaze entropy (SGE) was observed to decrease in the experimental group during these two phases compared to the control group, neither SGE nor gaze transition entropy (GTE) exhibited significant reductions (**Fig. 7g**, **7h**). The dual-monitor setup may have increased both spatial and transition entropy by requiring gaze shifts between the original and adjusted screens.

Furthermore, the number of verbal instructions issued to the camera assistant was lower in the experimental group during these two phases (**Fig. 7d**) (Calot: Mean difference = −2.92, Cohen's $dz = -1.20$, $p = 0.002$; Diss: Mean difference = −1.83, Cohen's $dz = -0.90$, $p = 0.019$), reflecting a reduced need for explicit coordination.

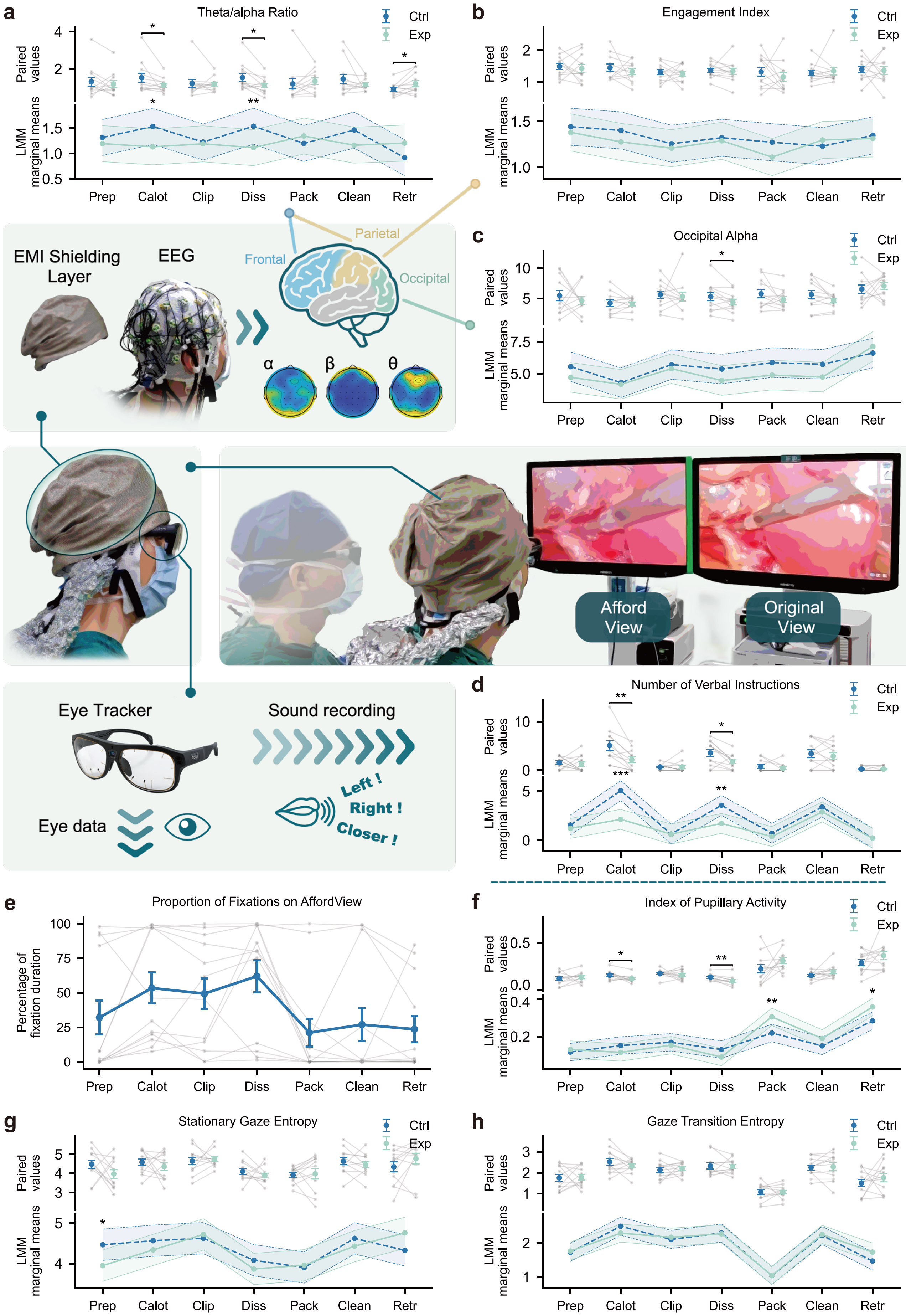

a
Theta/alpha Ratio
Ctrl
Exp
Paired values
LMM marginal means
Prep
Calot
Clip
Diss
Pack
Clean
Retr
b
Engagement Index
c
Occipital Alpha
EMI Shielding Layer
EEG
Frontal
Parietal
Occipital
α
β
θ
Afford View
Original View
Eye Tracker
Sound recording
Eye data
Left !
Right !
Closer !
d
Number of Verbal Instructions
e
Proportion of Fixations on AffordView
Percentage of fixation duration
f
Index of Pupillary Activity
g
Stationary Gaze Entropy
h
Gaze Transition Entropy

**Fig. 7 | Setup for the intraoperative human-factors study and objective metrics.** The central illustration depicts the experimental configuration deployed in the operating room during paired laparoscopic cholecystectomies. The setup features a dual-monitor system displaying both the original view and auto-framed view (AffordView). For the objective assessment of the application's effectiveness, the surgeon is equipped with an EEG cap incorporating EMI shielding layers to reduce radiofrequency noise from energy devices, and a wearable eye-tracker that simultaneously records verbal instructions. **a–c**, Intraoperative EEG metrics comparing the Ctrl and Exp conditions across discrete surgical phases, specifically the TAR (**a**) and EI (**b**), as well as occipital alpha band power (**c**). **d**, The number of verbal instructions issued to the camera assistant in the Ctrl versus Exp groups. **e**, Percentage of fixation duration on the AffordView monitor within the Exp condition across surgical phases. **f–h**, Comparisons between the Ctrl and Exp conditions for IPA (**f**), SGE (**g**), and GTE (**h**). *, $p < 0.05$; **, $p < 0.01$; ***, $p < 0.001$; EEG, electroencephalography; EMI, electromagnetic interference; TAR, theta/alpha ratio; EI, engagement index; Ctrl, control; Exp, experimental; Prep, Preparation; Calot, Calot triangle dissection; Clip, Clipping and cutting; Diss, Gallbladder dissection; Pack, Gallbladder packaging; Clean, Cleaning and coagulation; Retr, Gallbladder retraction; IPA, index of pupillary activity; SGE, stationary gaze entropy; GTE, gaze transition entropy.

**Subjective metrics:** The objective and behavioral findings were corroborated by the subjective assessments. Surgery Task Load Index (SURG-TLX)[46] scores were significantly lower in the experimental group (Mean paired difference = $-1.52$, Cohen's $dz = -0.83$, exact paired permutation $P = 0.014$; $n = 12$ pairs), with the Mental Demands and Physical Demands subscales primarily contributing to this reduction (**Fig. 8a**, **8b**). Surgeons' ratings of the application's utility also showed higher scores during the phases on which the model was trained (**Fig. 8c**).

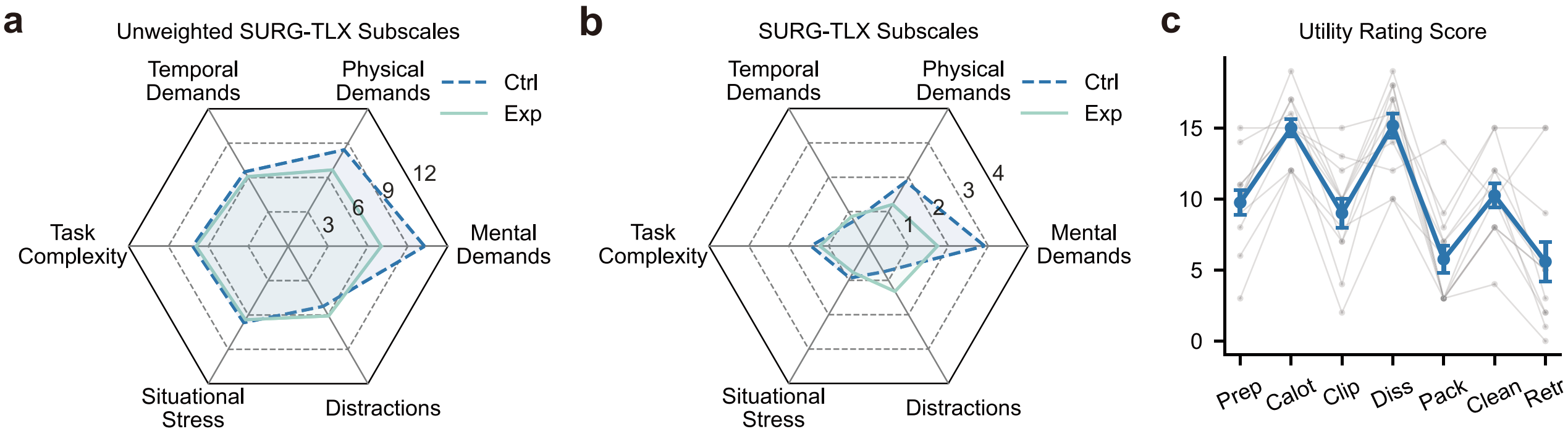

**Fig. 8 | Subjective metrics.** **a**, **b**, Cognitive workload assessed via the SURG-TLX, presented as radar charts detailing the unweighted (**a**) and weighted (**b**) subscale scores. **c**, Surgeons' intraoperative utility ratings of the AffordView application across distinct surgical phases. SURG-TLX, Surgery Task Load Index.

## Discussion

The proposed framework grounds tissue affordance retrospectively and, from these labels, learns to predict the spatial ROI from surgical context. Rather than passively estimating where the surgeon looks, this intention-aware modeling anticipates where the surgeon is about to act, bridging the conceptual gap between the reactive mechanics of explicit selection and the cognitive foresight of implicit prediction. Affordance grounding has also been extended from rigid objects[47] to deformable ones, but typically remains anchored to a stable part inventory: garment affordances are attached to named components such as a sleeve or a hem, so that one annotated exemplar transfers across instances of a category[17]. Soft tissue admits no comparable decomposition, since interaction loci follow procedural intent rather than object identity, and the same region may afford different actions at different moments. Supervision can alternatively be generated by acting, as in deformable-object manipulation, where candidate actions are scored against a formalized task objective using interaction the system collects itself[48]. Surgical view control admits neither route, because exploratory interaction cannot be performed on a patient and the operative goal resists formalization as a measurable target state. Within surgery, dense spatial guidance has instead been supplied normatively, by annotating where dissection is permissible rather than where it occurs: Go/No-Go models render expert-defined safe and unsafe zones as heatmaps and have been benchmarked against panels of expert surgeons[49]. Such maps prescribe where action is allowed, whereas the affordance grounded here describes where expert action in fact took place, so the two answer different questions about the same scene.

To address these complexities, the concept of tissue affordance has recently emerged. A closely related effort, AffordTissue, a concurrent preprint, likewise predicts dense tissue-affordance heatmaps anchored to tissue rather than to anatomical categories[50]. The two diverge less in machinery than in a prior question: where the spatial supervision for what affords interaction should come from, given that this intent is tacit and resists precise explicit articulation. AffordTissue supplies it externally, from expert per-frame annotations under a language specification of the surgery, tool, and action; our framework derives it internally, treating the surgeon's consummated trajectory as evidence for where interaction was afforded, without exhaustive manual labeling. Which source is appropriate follows from the downstream objective, in keeping with the relational nature of affordance[21], which is defined over the agent–environment system rather than as a fixed property of tissue[51]. An externally specified, verifiable target suits a robotic safety constraint, whereas a behaviorally derived one suits the modeling of a surgeon's anticipatory attention for proactive view control, a setting in which soliciting real-time language would itself add to the workload. The two thus address distinct objectives and are complementary rather than competing.

Within this formulation, affordance constitutes an epistemic prior rather than a static anatomical attribute:

the evidence that a region affords manipulation accumulates over the unfolding operation. Retroactively analyzing the operative trajectory infers interaction goals and grounds hotspots specific to the scene, defining targets through accumulated actionability rather than static anatomical landmarks. This actionability-based definition is supported by the close alignment of the generated hotspots with true intraoperative surgeon gaze.

To realize this retrospective grounding, DiffeoAfford models cross-frame correspondence with a diffeomorphic deformation field fitted jointly with a global transform. Prior natural-scene affordance grounding relies on a global transform alone, which captures dominant camera motion but has too few degrees of freedom for the non-rigid distortion of compliant tissue. Adding a diffeomorphic field supplies the deformation freedom while remaining smooth, invertible, and topology-preserving by construction, precluding the crossings and folds that would violate biological continuity. With the rest of the pipeline fixed, replacing this field with a global transform alone raised grounding error (**Supplementary Fig. 3**). Combined with the global transform handling macroscopic camera motion, it yields the dense, pixel-wise correspondence that attaches instrument tips to the tissue and transports them through deformation across a clip (**Fig. 2**). Conceptually, this aligns with recent surgical work that regularizes tissue deformation toward physical plausibility; Gong et al., for instance, use diffeomorphic constraints to recover non-crossing, realistic warps of manipulated tissue[52]. We operate directly on the 2D image plane, which the viewing geometry justifies and which avoids the cost and instability of dense 3D reconstruction (Supplementary Information).

Agreement between the grounded labels and human reference was established in the cholecystectomy evaluations. On Cholec80, the annotations were statistically equivalent to those of novice annotators given full procedural context (**Supplementary Fig. 2c**), and on LCET they aligned with intraoperative surgeon gaze (**Fig. 4d**). The pipeline also applied to hysterectomy video, where the grounded labels were sufficient to train an AutoLaparo-specific predictor. That experiment evaluated downstream prediction and camera-motion forecasting rather than grounding accuracy, so it supports the transferability of the grounding-to-prediction framework across procedures rather than grounding fidelity in a second procedure. The result is also consistent with the diffeomorphic prior encoding deformation behavior common to soft tissue rather than procedure-specific anatomy.

Beyond spatial coherence, the framework addresses the temporal dimension by aligning model output with the anticipatory behavior of the surgeon. Unlike foresight in deformable-object manipulation, which denotes a long-horizon value over the agent's own prospective actions[48], the future anticipated here belongs to a different agent, and affordance grounding outside surgery is typically static, resolving where to act in the current frame without a temporal lead[17]. Within surgery, spatial attention has more often been modeled by regressing observed gaze, for example by jointly predicting scanpath and instrument masks with the explicit aim of informing camera guidance[53]; deriving the target from action instead leaves gaze available as an independent criterion rather than as the training signal. Contemporary approaches model surgical temporal dynamics either by recognizing or anticipating surgical phases[54, 55] and action triplets[12], or by forecasting the timing of camera movements from instrument kinematics. The triplet formalism carries an explicit interaction target, but in public benchmarks it is typically a semantic class without spatial annotation, and any localization recovered as a weakly supervised by-product need not coincide with the true anatomical target[12]. Kinematics-based models anticipate when a camera move is imminent but not where the interaction lies[56]. DiffeoAfford instead grounds a dense, continuous AH from instrument–anatomy interactions, yielding a fine-grained spatial target; incorporating explicit action-level semantics into this substrate is a natural extension rather than a competing approach, as discussed below.

In contrast to purely reactive frameworks, the present approach acknowledges that expert surgeons organize interventions through proactive gaze behaviors, including target locking and quiet eye strategies. These behaviors fixate the interaction target before the hand arrives, and expert surgeons sustain markedly longer pre-movement fixations than novices[23, 57], satisfying the feedforward information requirements of the motor system. Training the prediction model on retrospectively aggregated interaction data enables it to estimate where interaction is expected to occur rather than to react to current tool contact. The LCET prediction evaluation examined this output against intraoperative attention directly. The centroid lag of AH prediction relative to surgeon gaze was significantly shorter than that of camera-assistant gaze in both evaluated phases, while group-mean profiles peaked on the leading side for AH prediction and on the lagging side for camera-assistant gaze (**Fig. 5a**). The peak lags themselves are descriptive, and no consistent temporal precedence over surgeon gaze was established. Spatially, AH prediction was significantly closer to surgeon gaze than either camera-assistant gaze or the image center (**Fig. 5b**). On AutoLaparo, the separately trained model reached 95.16% directional consistency immediately before camera motion and retained accuracy over longer horizons as the reactive instrument-tracking baseline degraded (**Fig. 5c–e**). The anticipation claimed here is therefore with respect to instrument action and subsequent camera motion, not with respect to surgeon gaze, against which the model shows close coupling with less lag than a human assistant.

This comparison has a practical reading. Because the assistant frames the view according to where they look, divergence between the assistant's gaze and the surgeon's attention is the point at which the surgeon must intervene verbally, which is consistent with the lower number of verbal instructions recorded in the separate deployment cohort (**Fig. 7d**). This likely reflects the structure of the assistant's task rather than individual competence: the assistant must infer the surgeon's intent from the same image while simultaneously operating the scope.

Having established that the framework anticipates the surgeon's target, we next ask why centering it reduces cognitive load. Minimally invasive surgery fundamentally alters the surgeon's perceptual-motor environment. The dissociation of visual and motor axes, loss of tactile input, and restricted two-dimensional endoscopic view create a tunnel vision effect that severely diminishes peripheral spatial awareness. Consequently, precise field-of-view (FoV) control acts as an essential cognitive scaffold, providing ergonomically optimized visualization to alleviate the substantial burden of navigating this restricted space. Originating from gaze-contingent displays that mimic human foveated vision[58], visualization technologies in surgical robotics have evolved into active camera control. Gaze-driven control, however, treats the eye as a joystick and is vulnerable to the Midas Touch problem; sustained directional control carries higher workload than embedded alternatives[59], and explicit object-based selection disrupts natural visual exploration. An alternative, implicit gaze-based attention estimation, as in GazeScope, avoids explicit selection but still requires the surgeon's gaze as input[60], motivating an action-derived route to view control.

AffordView requires simple hardware (a PC with monitor) and aligns the displayed field with the visuomotor constraints under which the surgeon operates. Conventional surgical monitors impose a rigid, two-dimensional spatial reference frame that triggers a central fixation bias, which acts as a robust attractor during scene exploration. This persistent bias acts as a spatial prior, influencing fixation behavior even when task-relevant visual saliency cues are present[61]. We observed this phenomenon in our clinical eye-tracking dataset, where the mean surgeon fixation location exhibited close proximity to the geometric center of the display (Mahalanobis distance of 0.05).

By proactively centering the predicted surgical interaction locus, AffordView leverages this natural viewing tendency to provide a stable visual reference. Unlike reactive instrument-tracking approaches or explicit gaze-contingent systems, which may require additional calibration or deliberate gaze control, our intention-aware framework maintains task-relevant information near a predictable high-priority coordinate. Consequently, the operator avoids expending top-down executive effort to override the inherent spatial gravity of the monitor. Intraoperative measurements support this account, showing significant reductions in the surgeons' theta/alpha ratio and index of pupillary activity during targeted surgical phases. Subjective workload assessments corroborated these neurophysiological outcomes, with surgeons reporting significant reductions in both mental and physical demands on the SURG-TLX scale. In multicenter surveys, surgeons preferred this proactive centering strategy to both the unadjusted view and a conventional instrument-tracking baseline.

By maintaining spatial congruence between the active surgical site and the display center, auto-framing reduces competition between task-relevant and display-centered reference frames[62]. This alignment may decrease the need for top-down attentional control to compensate for persistent spatial biases, allowing visual attention to remain more closely coupled with surgical objectives[62, 63]. Intraoperative eye-tracking data support this effect, showing increased fixation duration on the adjusted view during Calot triangle and gallbladder dissection phases, consistent with greater engagement with the task-relevant surgical field.

Several limitations remain. The current framework predicts tissue affordance from visual evidence but lacks semantic reasoning about why specific regions warrant interaction, and uncertain visual cues can present alternative plausible hotspots. For example, while a tissue plane may physically afford dissection, semantic awareness of gallbladder cancer infiltration is required to prevent interaction with malignant margins and avoid tumor seeding. This diverges fundamentally from the methodology of AffordTissue[50]. Whereas AffordTissue relies on language conditioning to explicitly instruct and generate tool-action specific affordance regions, we envision introducing VLMs (e.g., SurgVLM and EnVR-LPKG) strictly to inject this overarching semantic knowledge into our existing visually grounded framework[64, 65]. This task, however, constrains how language may enter: view control must not depend on frequent intraoperative input from the surgeon, which would reinstate the very cognitive burden the framework seeks to reduce. Since most VLMs require a prompt at inference, even if it can be minimal or hidden, how to supply this semantic prior with little or no real-time input from the surgeon is the central open question.

Furthermore, because the existing spatial centering function relies on digital cropping, it does not relieve the restricted peripheral awareness imposed by the narrow endoscopic view[27]. Cropping reduces the displayed field further. Holding the optics and scope position fixed did, however, isolate the effect of target selection from any confound of physical camera motion, since the control and experimental sessions differed only in where the display was centered. The resolution and field-of-view trade-off this reflects has motivated dual-view foveated laparoscopes that retain a wide overview alongside a magnified one, on the grounds that events outside the frame, including inadvertent contact by energized instruments, may go unrecognized[66]. A related caution applies to attentional guidance itself: augmented-reality navigation has been reported to improve targeting accuracy while reducing detection of unexpected findings that remain within the visual field[67], so the reductions in cognitive load observed here should not be read as evidence of preserved peripheral vigilance, which was not measured. A further constraint concerns the comparison group: the camera assistants in the deployment cohort were novices with fewer than five prior procedures, the population in which misreading of surgical intent is most frequent. Whether the observed reduction in workload persists with experienced assistants remains to be established. Although wide-angle endoscopy can provide partial FOV expansion, addressing this limitation may require integrating AffordView with a robotic endoscope capable of physically repositioning the camera, such as a kinematically redundant 7-degree-of-freedom manipulator. This could enable physically actuated camera control subject to remote-center-of-motion constraints while optimizing viewing angles, emulating the anticipatory competence of an expert camera operator. More broadly, because the affordance hotspot represents where interaction is expected rather than a view-control signal specifically, the same predictions could support other intraoperative functions, each of which would require its own validation.

## Materials and methods

### Datasets

**Cholec80 dataset[54]:** A public dataset containing 80 cholecystectomy videos, with surgical phases labeled, is widely used in laparoscopic surgery research. Each video was recorded at a frame rate of 25 fps, with a resolution of 854 × 480 pixels.

**AutoLaparo dataset[68]:** A public dataset comprises 21 laparoscopic hysterectomy videos, each recorded at 25 fps with a standard resolution of 1920×1080 pixels. Three sub-datasets are designed for the three tasks. For task 1, all 21 videos are annotated with surgical phases. For task 2, video clips are selected from Phases 2 to 4, each lasting 10 seconds, with camera motion starting at the fifth second, and was labeled according to its motion category.

**Private laparoscopic cholecystectomy eye-tracking (LCET) dataset:** Between July 2024 and June 2026, 26 elective laparoscopic cholecystectomies for benign gallbladder disease were prospectively collected at Wuhan Union Hospital using non-consecutive convenience sampling when research staff were available. Adults were eligible; cases involving other operations, conversion to open surgery, failed eye-tracker calibration, incomplete videos, or corrupted recordings were excluded. All patients provided written informed consent for video recording, and all participating surgical staff provided written informed consent before data collection; all data were de-identified before analysis. Surgeon gaze was recorded in all procedures using Tobii Pro Glasses 3 (100 Hz); 13 had surgeon-only gaze and 13 had synchronized camera-assistant gaze recorded using Tobii Pro Glasses 2 (50 Hz), for the grounding and prediction evaluations, respectively. Analyzed operative duration was defined as the cumulative duration of all annotated surgical phases excluding Preparation (median, 15.0 min; IQR, 12.1–23.3 min; Supplementary Figs. 7a and 8).

Surgeon and camera-assistant gaze recordings were synchronized with the laparoscopic video and mapped to a common video-frame coordinate system; acquisition and alignment procedures are detailed in the Supplementary Information (**Supplementary Fig. 5**).

### DiffeoAfford pipeline for affordance hotspot dataset generation

To retrospectively extract surgical AHs and construct a training dataset, we developed DiffeoAfford pipeline (**Fig. 2**). We observed that while novices struggle to predict the interaction locus prospectively, they can localize it in hindsight after viewing the complete operation. This phenomenon indicates that the definition of the target is inherent to the expert's execution, where the relevant affordance hotspot is revealed through the action itself[30]. Consequently, the surgeon's operation trajectory attached to the tissue can serve as action-derived supervision for these hotspots. This approach is consistent with prior egocentric-vision studies that infer actionable regions from human demonstration videos. By treating the laparoscopic view as a specialized class of egocentric data, we can leverage these data-driven pipelines to extract affordance labels from expert traces automatically. Given the lack of a universally accepted representation for affordance, prediction problems vary by task, ranging from object localization and functional classification to segmentation, end effector pose estimation, and synthesis[69].

For view control tasks that require spatial location to center the view, we treat functional segmentation as the target form of affordance prediction. DiffeoAfford depicted in **Fig. 2** is designed to retrospectively ground AH in laparoscopic videos.

To achieve this, the pipeline executes the following steps: (1) It first performs instance segmentation on the active surgical instruments to locate their tips and orientations (**Instrument Segmentation & Localization**). (2) Based on instrument motion, the video is initially divided into operative demonstration clips. The pipeline then iteratively applies the following sub-steps to subdivide these into single-action demonstration clips and generate corresponding AH labels (**Temporal Video Segmentation**): for each clip, it establishes dense tracking across frames to account for tissue displacement (**Spatial Correspondence**), aggregates past and future interactions onto a single reference frame and propagates them back to all other frames (**Aggregation** and **Propagation**) , and models these points into continuous labels (**Refinement**). Finally, this yields an AH dataset comprising video clips paired with AH annotations.

**Instrument Segmentation & Localization:** Initially, the active surgical instruments were segmented using the Segment Anything Model 2 (SAM2), chosen for its proven effectiveness in surgical scenarios[70], and the instrument tips and orientations were derived from the resulting segmentation mask for the subsequent steps.

**Temporal Video Segmentation:** Most existing affordance grounding methods rely on datasets comprising manually segmented single-action demonstration clips[31, 71, 72, 73]. However, applying this manual protocol to surgical videos would be prohibitively time-consuming and fundamentally contradict our goal of achieving an automated workflow. We therefore segmented surgical videos automatically based on instrument activity and AH visibility, with detailed in the Supplementary Information.

For each generated clip, the following steps are executed to generate the corresponding AH labels: **Spatial Correspondence:** Following prior work on affordance grounding in egocentric settings[30, 31, 74], frame-to-frame spatial correspondences are typically established via global transformations (e.g., homographic transforms). Therefore, advanced point trackers robust to occlusions are employed to obtain sparse spatial correspondences[75]. DiffeoAfford combined these sparse correspondences with global transforms and local diffeomorphic deformation; query-frame selection, aggregation, propagation, and refinement are detailed in the Supplementary Information.

Similarity- and homography-RANSAC baselines used the same tracking points and downstream processing as DiffeoAfford, differing only in the correspondence model; implementation details are provided in the Supplementary Information.

**Real-time affordance hotspot prediction model**

To translate offline annotations into proactive intraoperative assistance through real-time prediction, we trained the SegFormer model on the datasets generated by DiffeoAfford (**Fig. 2**). SegFormer is a simple and efficient transformer-based semantic segmentation architecture[76], which has proven effective for laparoscopic tissue recognition[77,78]. To adapt the model to our specific task, the output layer was modified to generate a single-channel affordance heatmap. To proactively predict targets without instrument-tracking bias, we trained exclusively on frames in the first half of each clip where the AH-to-instrument distance exceeded the average distance. During prediction, the centroid of the generated heatmap serves as the predicted AH location. SegFormer-B3 was used and achieved an inference speed of 70 FPS on an NVIDIA RTX 4090 GPU.

**AffordView application**

AffordView centers the displayed FoV on the predicted AH, crops the frame while preserving its aspect ratio, and upscales it to the original display resolution, with boundary constraints limiting magnification to 1.67×; detailed equations are provided in the Supplementary Information.

To deploy this application for our clinical evaluations, a specific instance of the prediction model was trained. DiffeoAfford generated 542 valid clips from the Calot triangle and gallbladder dissection phases of Cholec80 videos 01–40. Following the original video order, the dataset was split at the video level into training (videos 01–32; 423 clips), validation (videos 33–36; 49 clips), and test (videos 37–40; 70 clips) sets. On the test set, the SegFormer-B3 model achieved an accuracy of 45 ± 30 pixels (measured via Euclidean distance). Based on feedback from the initial surgeon preference evaluation, we also incorporated a smart zoom control mechanism to enhance the user experience. We observed that zoom adjustments are primarily driven by instrument dynamics rather than spatial target positioning. Therefore, we integrated a timing inference method based on instrument motion characteristics (developed in our previous work[79]) to determine the moments for zooming in and out. Implementation details are provided in the Supplementary Information.

Together, these components convert retrospective data annotation into proactive surgical assistance. DiffeoAfford generates dense tissue-affordance labels from surgical videos. A real-time model learns from these labels to predict interaction hotspots. AffordView then uses these predictions, together with instrument-derived zoom cues, to adjust the displayed FoV during live procedures.

**Evaluation setup**

**Evaluation of DiffeoAfford on Cholec80 dataset:** We evaluated the annotation accuracy of DiffeoAfford using the Calot triangle dissection phase from videos 01–10. We selected the tip of the electrosurgical hook as the primary instrument reference point. The pipeline yielded 141 video clips for evaluation. To reduce instrument-induced annotator bias, we randomly sampled one frame per clip where the distance between the pipeline-annotated AH center and the instrument tip exceeded the dataset's mean distance. Three annotator groups were recruited: senior surgeons ($n = 3$, >10 years' experience, serving as ground truth), junior doctors ($n = 3$, 2–3 years' experience), and medical students ($n = 3$, novices). Senior surgeons annotated images after viewing the corresponding video clips. Junior and novice groups initially annotated static images without context; the novice group then re-annotated them after viewing the contextual clips (In-context).

**Evaluation of DiffeoAfford against intraoperative gaze on LCET dataset:** This evaluation used the 13 LCET procedures with surgeon gaze recordings only. The pipeline generated 60 video clips and randomly sampled one frame every 6 seconds, yielding 164 images. Unlike the Cholec80 evaluation, the surgeon's actual intraoperative gaze points served as the proxy ground truth. Let $T_s$ be the timestamp of the sampled image. Three temporal ground truths were established by calculating the mean coordinates of gaze points within the windows $T_s \pm 0.5$s, $T_s \pm 1.0$s, and $T_s \pm 2.0$s. We computed the Euclidean distances from these ground truths to the AH annotations provided by both the pipeline and the human annotators ($n = 3$ junior doctors, before and after watching the 10-second reference clips).

**Evaluation of AH prediction against intraoperative gaze on the LCET dataset:** This evaluation used the 13 LCET procedures with synchronized surgeon and camera-assistant eye tracking, restricted to the Calot triangle dissection and gallbladder dissection phases. AH prediction sequences were generated using the same Cholec80-trained SegFormer-B3 model used in the AffordView application, taking the framewise centroid of each predicted heatmap. All gaze coordinates were mapped to the 1920 × 1080 laparoscopic-video coordinate system; surgeon and camera-assistant gaze were placed on a common 50-Hz time grid, and timestamped AH predictions were assigned to the nearest grid samples. Non-contiguous intervals belonging to the same phase were retained as separate segments so that temporal pairs were not formed across segment boundaries.

Temporal alignment was evaluated using cross-recurrence quantification analysis (CRQA), which has previously been used in dual eye-tracking studies to characterize temporal coordination and leader-follower relationships between paired gaze sequences[80]. Because the laparoscopic field of view changes continuously, the analysis was restricted to a diagonal band around the main diagonal, which focused the analysis on temporally local recurrence within the same evolving surgical scene[80], and was summarized by the diagonal cross-recurrence profile (DCRP), which describes the direction and magnitude of temporal coupling between two sequences[81]. The band spanned candidate lags from −3.0 to +3.0 s in 20 ms increments, a window selected from previous gaze cross-recurrence research focusing on recurrence close to the main diagonal and the approximately 3 s intervals between camera movements reported for experienced surgeons during camera targeting tasks[80, 82]. Recurrence-rate and centroid-lag definitions are provided in the Supplementary Information.

A negative candidate lag indicates that the candidate sequence leads surgeon gaze. Procedure-level recurrence-rate profiles were averaged with equal weight across procedures to obtain the displayed group profile, and the shaded bands were calculated as the mean plus or minus 1.96 standard errors. For each procedure, the DCRP was treated as a distribution over candidate lags, and temporal alignment was summarized by its recurrence-rate-weighted mean, referred to here as the centroid lag, following the lag-profile distribution approach described in previous visual-attention research[83]. The centroid-lag equation and interpretation are provided in the Supplementary Information.

Spatial alignment was assessed at simultaneous valid timestamps using procedure-level median distances from surgeon gaze; a positive median advantage indicates that AH prediction was closer than the comparator. Detailed sampling and formulae are provided in the Supplementary Information.

**Evaluation of real-time AH prediction for camera motion forecasting on AutoLaparo dataset:** To investigate the relationship between the model-predicted AH and actual camera motion, we evaluated our method on the AutoLaparo dataset. Because the standardized video clips for AutoLaparo Task 2 (which evaluates camera motion prediction) were exclusively extracted from phase 2 (Dividing Ligament and Peritoneum), phase 3 (Dividing Uterine Vessels and Ligament), and phase 4 (Transecting the Vagina), we specifically targeted these three phases for our evaluation. Using the tips of the bipolar forceps and electric hook as primary instruments, DiffeoAfford generated affordance annotations for 435 clips from Task 1 videos 01–10. Following the original video order, the dataset was split at the video level into training (videos 01–08; 346 clips) and validation (videos 09–10; 89 clips) sets for an AutoLaparo-specific SegFormer-B3 model. A total of 124 AutoLaparo Task 2 video clips involved four camera motion directions: up, down, left, and right. The model was used to predict AH locations in these clips. The final prediction result was defined as the temporal mean of the AH locations predicted within a 0.5-second window prior to the camera motion event. This was compared against the **instrument-tracking baseline** (detailed in the **Supplementary Information**), which used the same 0.5-second temporal mean protocol.

**Surgeon preference assessment of AffordView application:** We designed two paired comparisons via an online questionnaire: (1) Original video vs. AffordView application using proposed AH prediction model, and (2) AffordView using proposed AH prediction model vs. the instrument-tracking baseline. To maximize the observable differences for evaluation, we selected 30 thirty-second clips for each comparison from the Calot triangle dissection phase of videos 41–50 in the Cholec80 dataset (extracting 3 clips per video). The selection criterion was based on the maximum average spatial distance between the target points generated by the two respective methods within that clip. Notably, for the original video, the geometric center of the frame was designated as its target point. Three versions of the questionnaire were deployed via the Python Flask framework (**Supplementary Fig. 7**), each containing 20 randomly ordered pairs (10 from each comparison).

**Intraoperative human-factors study of the AffordView application:** This prospective, non-randomized, paired, within-subject human-factors study was conducted from 14 July to 25 August 2025 to evaluate surgeon–system interaction and cognitive workload during intraoperative use of AffordView. Cases were enrolled by non-consecutive convenience sampling when research staff were available. Eligible patients were adults undergoing elective laparoscopic cholecystectomy for benign gallbladder disease; cases involving other operations, conversion to open surgery, failed eye-tracker calibration, incomplete videos, or corrupted recordings were excluded. All patients provided written informed consent for video recording, and all participating surgical staff provided written informed consent before data collection; all data were de-identified before analysis. The study recruited 5 operating surgeons and 12 novice camera assistants (each with experience in fewer than 5 procedures). The sample size was determined a priori using a power analysis for paired comparisons (two-sided $\alpha = 0.05$, power = 0.80, expected Cohen's $dz = 1.0$ based on effect sizes reported in prior surgical cognitive workload studies[84]), yielding a minimum requirement of 10 pairs. Twenty-four laparoscopic cholecystectomies were performed and organized into 12 matched pairs to evaluate the application's overall ergonomic impact. Each pair consisted of a control session (dual monitors only, original view only) and an experimental session (one original view monitor, one adjusted view monitor). The same surgeon and camera assistant participated in both sessions within each pair, and condition order was assigned by strict alternation rather than randomization to balance sequence effects. Analyzed operative duration was defined as the cumulative duration of all annotated surgical phases, including Preparation (Supplementary Figs. 7b and 9). This paired within-subject design, in which each surgeon was assessed under both viewing conditions with concurrent eye tracking and cognitive-load measures, was adapted from Anschuetz et al.[85] This evaluation incorporated subjective CWL assessment via the SURG-TLX questionnaire, alongside objective EEG and eye-tracking metrics. Details regarding EEG data acquisition and preprocessing are provided in the **EEG data acquisition and preprocessing** section. Eye-tracking data were collected using Tobii Pro Glasses 3 and processed with Tobii Pro Lab and Python; the analyses included IPA, fixation-duration proportions, SGE, and GTE, with metric definitions provided in the Supplementary Information.

Finally, the surgeon's ratings of utility and the number of verbal instructions to the camera assistant were recorded per surgical phase.

### EEG data acquisition and preprocessing

EEG was acquired with a 64-channel Brain Products LiveAmp system and processed in EEGLAB. Signals were filtered to 4–30 Hz, cleaned using ASR and ICA-based artifact removal, re-referenced, and summarized as phase-specific theta-, alpha-, and beta-band power; full preprocessing parameters are provided in the Supplementary Information.

### Statistical analysis

To control for confounding effects during the dataset evaluations, we employed linear mixed models (LMMs). For the Cholec80 dataset, the LMM included group as a fixed effect, with image and annotator as random intercepts. The novice (In-context) group served as the reference level to compute estimated differences against other groups. Similarly, for the LCET dataset, we applied the identical LMM structure, utilizing the junior (In-context) group as the reference level. To account for multiple comparisons against the reference level within these models, all resulting p values were adjusted using the Benjamini-Hochberg procedure.

Equivalence between DiffeoAfford and context-informed annotators was assessed using an image-level median bootstrap with 20,000 resamples and a data-driven margin based on within-human disagreement; equivalence was concluded when the 90% CI lay within −1 to 1. Global-transform baselines were compared using two-sided paired Wilcoxon signed-rank tests. Detailed definitions are provided in the Supplementary Information.

For the LCET prediction evaluation, each procedure was treated as the statistical unit, and framewise samples were used only to derive procedure-level metrics rather than as independent observations. Within each surgical phase, procedure-level DCRP centroid lags and median distances to surgeon gaze were compared using exact two-sided paired Wilcoxon signed-rank tests: AH prediction versus camera-assistant gaze for both metrics, and AH prediction versus the image center for distance. The p values of these six prespecified comparisons were adjusted jointly using the Holm procedure. Group-mean DCRP peak lags were treated as descriptive and were not subjected to statistical testing.

Surgeon-preference proportions were evaluated using exact binomial tests. In the intraoperative human-factors study, primary inference used two-sided paired permutation tests; LMMs were used as complementary analyses.

### Ethical approval

The study protocol was approved by the Medical Ethics Committee of Union Hospital, Tongji Medical College, Huazhong University of Science and Technology (UHCT250178). All participants in both the prospective LCET data-collection study and the prospective intraoperative human-factors study provided written informed consent before participation. Participants in the intraoperative human-factors study received no financial compensation.

## Acknowledgements

H.D., X.C., Y.W., C.W. and H.Z. disclose support for the research of this work from the Hubei Science and Technology Major Program [grant number 2023BCA002] and the National Key Research and Development Program of China [grant number 2022YFC2407402].

## Author contributions

J.G., X.C., Y.W., J.P. and H.D. conceptualized the project. H.D., X.C., Y.W., C.W. and H.Z. acquired the funding. J.G. designed and developed DiffeoAfford. J.G., J.Z., S.Z., H.S., Q.Y., Q.X. and X.C. designed and developed the AffordView application. J.G. and Y.W. designed the evaluation experiments for grounding and prediction of AHs, while J.G., J.P., X.C. and Q.F. designed the assessment experiments for the AffordView application. J.P. and Q.F. provided the eye tracker and technical support. W.C. and C.X. provided the electroencephalography equipment and technical support. G.C., K.L., Y.C., H.W. and S.S. performed the assessment experiments. J.G., J.Z. and H.W. analyzed the data. J.G., S.Z. and H.W. constructed the dataset. J.G., X.C., Y.W., J.P. and H.D. wrote and edited the paper.

## Competing interests

H.S., Q.Y. and Q.X. are employees of Wuhan United Imaging Surgical Co., Ltd., which provided engineering support for the development of the AffordView application; their contributions are detailed in the author contributions statement. The remaining authors declare no competing interests.

## Supplementary Methods

**Intraoperative eye-tracking acquisition and alignment:** To overcome the distance limitations that render standard screen-based eye trackers ineffective during laparoscopic surgeries, we developed a robust framework for intraoperative eye-tracking acquisition and processing. This framework enables synchronized data collection from multiple subjects and ensures precise temporal and spatial alignment of gaze points with the scene and laparoscopic videos. Through this alignment, the gaze data are mapped onto the video frame coordinate system (**Supplementary Fig. 5**). The same framework was applied to the surgeon and camera-assistant recordings, placing both gaze sequences and the framewise AH predictions on a shared video timeline.

**Temporal video segmentation:** Because surgical intent and interaction hotspots are intrinsically linked to instrument kinematics, these continuous actions can serve as a reliable proxy for manual segmentation boundaries. To capitalize on this, we designed an iterative refinement process that automatically proposes and refines candidate action segments by leveraging instrument presence and the visibility of the AH. Initially, the video is segmented into operative demonstration clips based on the visible length and duration of the active instrument, and corresponding AH annotations are generated for each clip. To refine these into single-action demonstration clips, the pipeline iteratively applies an automated subdivision process. Specifically, any clip is subdivided if it meets either of the following criteria: (1) over 50% of the AH area (defined as the region with heatmap intensity > 0.1) is occluded for longer than the occluded threshold $T_o$, or (2) the total clip duration exceeds the maximum length limit $T_{max}$. Affordance heatmaps are then regenerated for each clip. This subdivision process repeats iteratively until no clips meet the criteria, after which any remaining clips shorter than the minimum duration limit $T_{min}$ are discarded. Because continuous surgical maneuvers in the datasets typically last between 10 and 30 seconds, we defined our temporal parameters as $T_{min} = 10s$, $T_{max} = 30s$, and $T_o = 5s$. Through these settings, the pipeline automatically generates the candidate affordance datasets.

**Spatial correspondence, aggregation, propagation, and refinement:** Subsequently, global transforms are fitted to these sparse matches to produce dense spatial correspondences. However, prior work typically targets rigid, natural scenes[1, 2, 3], limiting their applicability in highly deformable anatomical environments. To address this, we propose a specialized method by incorporating diffeomorphic deformation fields alongside the existing global transforms (**Fig. 2 Spatial Correspondence**). This diffeomorphic constrained tracking allows better modeling of tissue deformation and reduces tracking noise (see Displacement estimation in Supplementary Information). Within each video clip, we select a query frame to serve as the tracking start frame. To optimize tracking performance by the trackable area and minimizing the fitted displacement, we select a frame from the middle third of the clip with the least instrument occlusion (operationalized as the shortest visible instrument length) as the query frame. The tracking region is defined as the entire frame region excluding the black border and the instruments. Using this setup, we obtain pixel-wise spatial correspondences between the query frame and the other frames in the clip. **Aggregation:** To aggregate the instrument tip locations, points from each frame are mapped to their corresponding coordinates in the query frame using the displacement field of that specific frame (**Fig. 2 Aggregation**). **Propagation:** Once aggregated, these points are propagated back to all other frames via reverse mapping, relying on the corresponding displacement fields (**Fig. 2 Propagation**). **Refinement:** To ensure accuracy on each frame, the mapped points are filtered using the Mahalanobis distance to eliminate outliers. The remaining inlier points are then modeled using a 2D Gaussian distribution. Specifically, a probability density function is fitted to the points and normalized to generate the final affordance label $A(x,y) = \exp\left(-\frac{1}{2}(\boldsymbol{p}-\boldsymbol{u})^T\boldsymbol{\Sigma}^{-1}(\boldsymbol{p}-\boldsymbol{u})\right)$, where $\boldsymbol{p}$ is the pixel coordinate vector, $\boldsymbol{u}$ is the mean vector of the filtered points, and $\boldsymbol{\Sigma}$ is their covariance matrix.

**Displacement estimation:** For each frame, a smooth, dense mapping is estimated to map pixels back to the query frame. This mapping is represented as a diffeomorphic deformation obtained by integrating a 2-D velocity field $\boldsymbol{v}$:

$$\phi(\boldsymbol{x}) = Integrate(\boldsymbol{v})(\boldsymbol{x})$$

and a configurable global transform $S$ (implemented as either a similarity or a homographic transform) is optimized jointly. Taking the similarity transform as an example, it is formulated as:

$$S(\boldsymbol{x}) = s \cdot \boldsymbol{R}(\theta)\boldsymbol{x} + \boldsymbol{t}$$

where $v$ denotes the velocity field, $s$ is an isotropic scale factor, $\theta$ is the rotation angle, and $\boldsymbol{t}$ is a translation vector. The deformation follows a Demons-style procedure[4]: at each optimization iteration, the velocity field $\boldsymbol{v}$ is smoothed by a Gaussian filter and then integrated by repeated composition to yield a diffeomorphic sampling grid $\phi(\boldsymbol{x})$. Finally, the parameters are optimized by minimizing an energy function $E$ that measures the average positional error of the sparse tracked points:

$$E = \min_{\boldsymbol{v},s,\theta,\boldsymbol{t}} \frac{1}{N} \sum_{i=1}^{N} \left\| S\left(\phi\left(\boldsymbol{x}_i^{(p)}\right)\right) - \boldsymbol{x}_i^{(q)} \right\|_2$$

where $\boldsymbol{x}_i^{(p)}$ represents the $i$-th tracked point in frame $p$, $\boldsymbol{x}_i^{(q)}$ is its corresponding location in the selected query frame, and $N$ is the total number of tracked points.

**Global-transform baselines:** Following prior natural-scene affordance-grounding methods[1, 3], we implemented two global-only correspondence models. Similarity and homography transforms were fitted between the query frame and each target frame using RANSAC with the same sparse tracking points used by DiffeoAfford. Instrument tips were then aggregated in the query frame and propagated to all frames using the fitted transforms. The same Mahalanobis-distance filtering (<2) and centroid calculation were subsequently applied. All other pipeline components were unchanged, thereby isolating the contribution of the diffeomorphic deformation field.

**AffordView field-of-view adjustment:** To utilize the real-time AH prediction model for auto-framing, we developed AffordView application (**Fig. 2**). The adjustments follow these specific protocols: **Target Centering:** The target position $\boldsymbol{c}_t = (x_t, y_t)$, determined either by the instrument-tracking baseline or by our proposed real-time AH prediction model, serves as the recommended view center. **Cropping and Upscaling:** The software preserves the original aspect ratio while cropping a region of width $w$ and height $h$ around $\boldsymbol{c}_t$. This cropped region is then upscaled to the original frame size $W \times H$ to produce the final adjusted FoV. **Boundary Constraints:** To prevent excessive cropping and magnification when the $\boldsymbol{c}_t$ is close to the image border, the crop dimensions are bounded such that $w \geq 0.6W$ and $h \geq 0.6H$. This establishes a maximum magnification limit of $1.67 \times$.

**Instrument-driven zoom control:** By learning from expert demonstrations, this module automatically detects keyframes in the video stream to trigger FoV adjustments. During a zoom-in event, the image is cropped and magnified around the target point (capped at 1.67× magnification); during a zoom-out event, this magnification effect is reverted.

**Temporal alignment analysis:** Let S(t) denote surgeon gaze and C(t) denote either AH prediction or camera assistant gaze. Recurrence at candidate lag τ was defined when the angular separation between surgeon gaze at time t and the candidate sequence at time t + τ did not exceed 1.5°, following the spatial proximity criterion used in previous gaze CRQA research[5], and the recurrence rate at each lag was calculated as the number of recurrent pairs divided by the total number of valid within-segment pairs at

that lag:

$$R(t,\tau) = I\big(d_\theta\big(S(t), C(t+\tau)\big) \leq 1.5°\big)$$

$$RR(\tau) = \frac{\sum_t R(t,\tau)}{N_{(valid)(\tau)}}$$

$$\tau_c = \frac{\sum_{\tau=-T}^{T} \tau \, \mathrm{RR}(\tau)}{\sum_{\tau=-T}^{T} \mathrm{RR}(\tau)}$$

where $\tau_c$ is the centroid lag, $RR(\tau)$ is the recurrence rate at candidate lag $\tau$, and $T = 3.0\ s$. Negative centroid-lag values indicate that the candidate tends to lead surgeon gaze, positive values indicate that surgeon gaze tends to lead the candidate, and values near zero indicate temporal coupling without a consistent lead or lag.

**Supplementary Fig. 6** illustrates this calculation. CRQA first represents recurrent surgeon–candidate locations across paired time indices as a cross-recurrence plot. Recurrences sharing the same temporal offset lie on a common diagonal; aggregating each diagonal produces the DCRP[6]. Centroid lag is the recurrence-rate-weighted center of this entire profile rather than the location of a single peak, thereby summarizing the overall direction and magnitude of temporal coupling[7].

**Spatial alignment analysis:** Spatial alignment was assessed at simultaneous zero-lag samples. Timestamps were retained when surgeon gaze, AH prediction, and camera-assistant gaze were all valid for the camera-assistant comparison, and when surgeon gaze and AH prediction were valid for the image-center comparison, with the image center defined as $(960,540)$ pixels. Euclidean distances to surgeon gaze were calculated at each retained timestamp and reduced to a median for each procedure and surgical phase. Median advantage was defined as the median across procedures of $d_{comparator} - d_{AH}$, so that positive values indicate a smaller median distance for AH prediction.

**Human-factors outcome definitions:** To evaluate the surgeons' cognitive workload, we calculated the theta/alpha ratio (TAR), Engagement index (EI) and occipital alpha band power (OA) as:

$$TAR = \frac{P_{\text{Frontal}\ \theta}}{P_{\text{Parietal}\ \alpha}}$$

$$EI = \frac{P_{\text{Parietal}\ \beta}}{P_{\text{Parietal}\ \alpha} + P_{\text{Parietal}\ \theta}}$$

$$\mathrm{OA} = P_{Occipital\ \alpha}$$

where $P$ represents the spectral power of a specific frequency band. The cortical regions were defined using the following electrode montages: F3, F4, and Fz for the frontal region; P7 and P8 for the parietal region; and O1 and O2 for the occipital region.

**Eye-tracking metrics:** Eye-tracking data were collected using Tobii Pro Glasses 3 and processed with Tobii Pro Lab and Python. Metrics included the index of pupillary activity (IPA), fixation duration proportions, and spatial entropy. For gaze entropy, the image area was discretized into 20×20 bins. Stationary gaze entropy (SGE) and gaze transition entropy (GTE) were calculated as:

$$SGE = -\sum_i p_i \log p_i$$

$$GTE = -\sum_i p_i \sum_j p_{ij} \log p_{ij}$$

where $p_i$ is the positional distribution probability in the $i$-th bin, and $p_{ij}$ is the transition probability from the $i$-th to the $j$-th bin.

**EEG data acquisition and preprocessing:** EEG data were acquired using a 64-channel Brain Products LiveAmp system, which was electromagnetically shielded to minimize radio frequency interference from energy devices, and processed using the MATLAB EEGLAB toolbox. Prior to workload analysis, continuous EEG signals underwent a standardized preprocessing pipeline. Non-EEG auxiliary channels (i.e., IMU sensors) were removed, and the data were bandpass filtered between 4 Hz and 30 Hz. Bad channels and high-amplitude artifacts were detected and rejected using the EEGLAB clean_rawdata function incorporating Artifact Subspace Reconstruction (ASR) with a burst criterion of 20. Missing channels were subsequently spherically interpolated, and the data were re-referenced to the common average. To eliminate ocular and myogenic artifacts, extended Infomax Independent Component Analysis (ICA) was applied. The EEGLAB ICLabel function was utilized to automatically identify and remove non-brain independent components with an artifact classification probability of ⩾ 90%. Following artifact rejection, the spectral power for the theta (4–8 Hz), alpha (8–12 Hz), and beta (12–30 Hz) frequency bands was calculated across the predefined surgical phases.

**Bootstrap equivalence analysis:** To assess whether DiffeoAfford achieved accuracy equivalent to that of context-informed human annotators, we conducted the same median-based, image-level bootstrap equivalence analysis for Cholec80 and LCET. For each image $i$, the human reference error $h_i$ was defined as the median error across the three Novice (In-context) annotators for Cholec80 or the three Junior (In-context) annotators for LCET, and the paired difference was calculated as $d_i = e_{DiffeoAfford,i} - h_i$. Within-human disagreement was defined as $q_i = median_{j<k}|e_{ij} - e_{ik}|$, and the data-driven equivalence margin as $\Delta = median_i(q_i)$, so equivalence denotes a discrepancy no larger than the typical disagreement between two human annotators. The standardized effect was calculated as $R = median_i(d_i)/\Delta$, a median-based paired statistic whose sign may differ from the LMM mean estimate. Images were resampled with replacement 20,000 times, with both $median_i(d_i)$ and $\Delta$ recalculated in each replicate. Equivalence was concluded when the 90% percentile bootstrap CI for $R$, corresponding to two one-sided tests at $\alpha = 0.05$, lay entirely within the interval from $-1$ to 1. Global-transform baselines were compared with DiffeoAfford using two-sided paired Wilcoxon signed-rank tests.

**Surgeon-preference and human-factors statistics:** For the surgeon preference assessment, preference proportions were evaluated using exact binomial tests against a null preference probability of 0.5. For the intraoperative human-factors study, permutation-based paired tests served as the primary inference method. The test statistic was the mean within-pair difference (experimental minus control), and the exact null distribution was obtained by enumerating all 2^12 possible sign flips. LMMs were subsequently used to further control confounding effects, incorporating condition, phase, and their interaction (condition × phase) as fixed effects, order as a fixed covariate, and pair as a random intercept. LMM inference used asymptotic Wald z tests; finite denominator degrees of freedom are not defined. All statistical tests were two-sided. Cohen's dz was calculated as the mean within-pair difference (experimental minus control) divided by the standard deviation of the within-pair differences. For forced-choice comparisons, odds ratios were calculated as the number of responses favoring the proposed method divided by the number favoring the comparator.

**Instrument-tracking baseline:** To establish a baseline for view control, we implemented the instrument-tracking-based target prediction method. Based on previous researches[8, 9, 10, 11], the center of multiple instrument tips is commonly used as the target location. Therefore, let $P_i$ denote the coordinate of the $i$-th detected instrument tip, the target center $C_{base}$ is defined dynamically based on the number of active instruments $n$ (**Fig. 2**): for a single instrument ($n = 1$), $C_{base} = P_1$; for two instruments ($n = 2$), $C_{base}$ is the midpoint $\frac{P_1+P_2}{2}$; and for $n \geq 3$, $C_{base}$ is the centroid of the polygon formed by all tips.

**Validity of the two-dimensional diffeomorphic approximation:** Intraoperative soft-tissue motion is a three-dimensional deformation $\Phi: S \rightarrow \mathbb{R}^3$ of the visible surface, whereas the pipeline estimates only its image-plane projection $\psi = \pi \circ \Phi$. Modeling $\psi$ as a two-dimensional diffeomorphism is a deliberate approximation justified by the operative viewing geometry rather than by full 3D reconstruction.

Decompose the surface displacement $u(x) = \Phi(x) - x$ into a component parallel to the image plane and a component along the optical axis, $\|u\|^2 = \|u \parallel\|^2 + \|u \perp\|^2$. Only the out-of-plane component $u \perp$ can create the depth-ordering reversals (self-occlusions) that would break a planar model; when $\|u \perp\|$ is small relative to $\|u \parallel\|$, the projected motion is well described by a smooth, invertible 2D map. The operative viewing geometry keeps $u \perp$ small. The in-plane component $u \parallel$ is the portion of tissue motion actually visible in the image, whereas $u \perp$ is largely invisible. To work under direct vision and maximize the manipulability and visibility of the target tissue, the surgeon orients the scope so that the relevant deformation unfolds within the image plane, maximizing $\|u \parallel\|$ and hence suppressing $\|u \perp\|$. The same clinical practice that makes a maneuver observable therefore suppresses the deformation component that would break the two-dimensional approximation.

The argument is applied per contiguous operative segment. Within one maneuver the viewpoint and depth ordering are stable, so $u \perp$ remains small. Modeling $\psi$ as a two-dimensional diffeomorphism is consistent with both the imaging geometry and the operative workflow, while avoiding the cost and static-scene assumptions of explicit three-dimensional reconstruction.

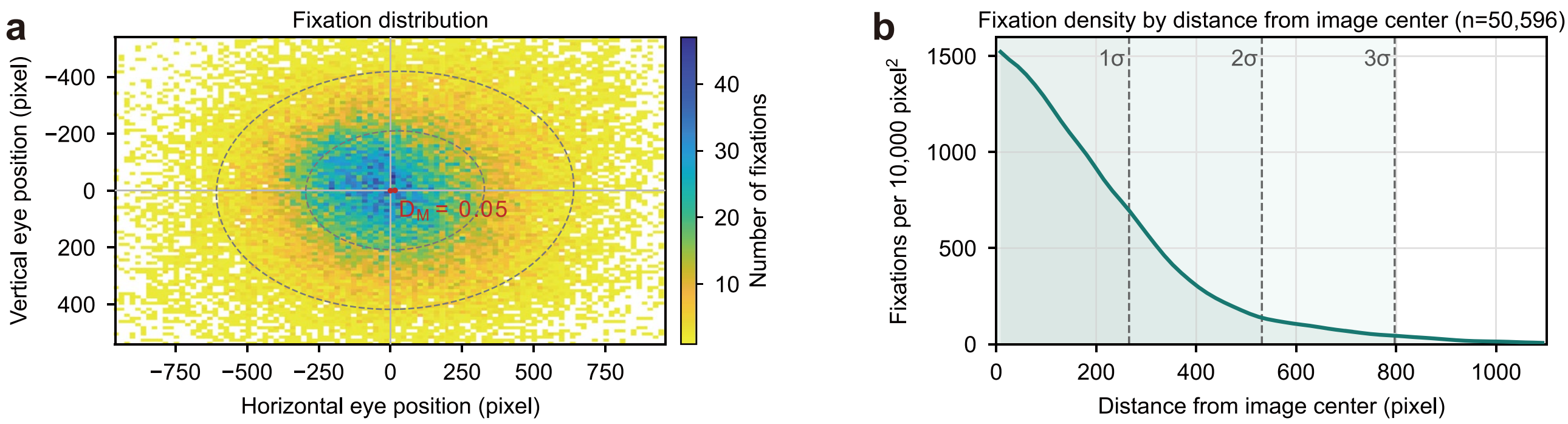

**Supplementary Fig. 1 | Center bias in laparoscopic surgery.** Analysis of 50,596 fixation centroids from 26 LCET procedures revealed a distinct center bias. **a**, Fixation distribution heatmap. Gray ellipses represent Mahalanobis distances of 1 and 2; the red line connects the mean fixation point to the geometric center of the monitor ($D_M = 0.05$). **b**, Area-normalized fixation density decreased with distance from the monitor center. Shading and dashed lines denote the $1\sigma$, $2\sigma$, and $3\sigma$ radial ranges ($\sigma = 265.7$ pixels).

**a**

All Groups vs. Novice (In-context)

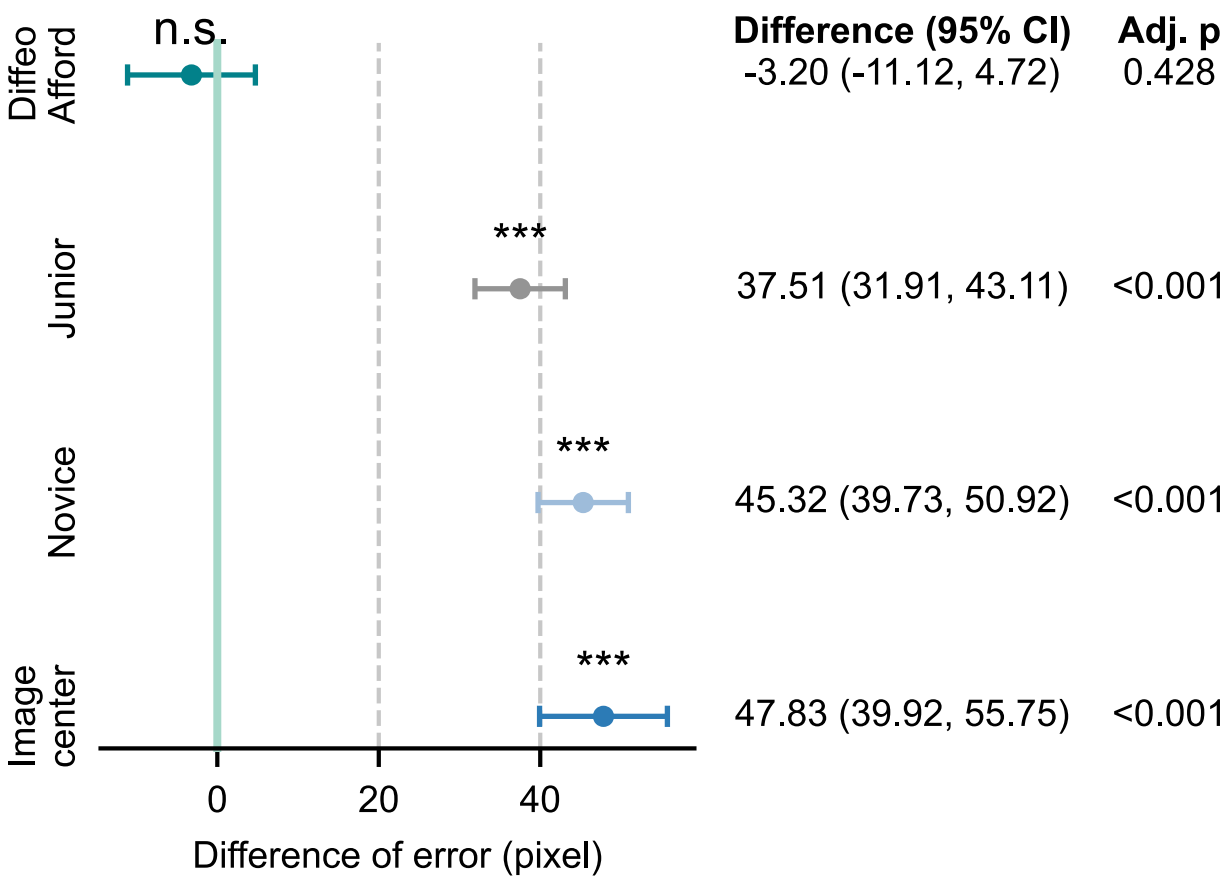

Residual Diagnostics

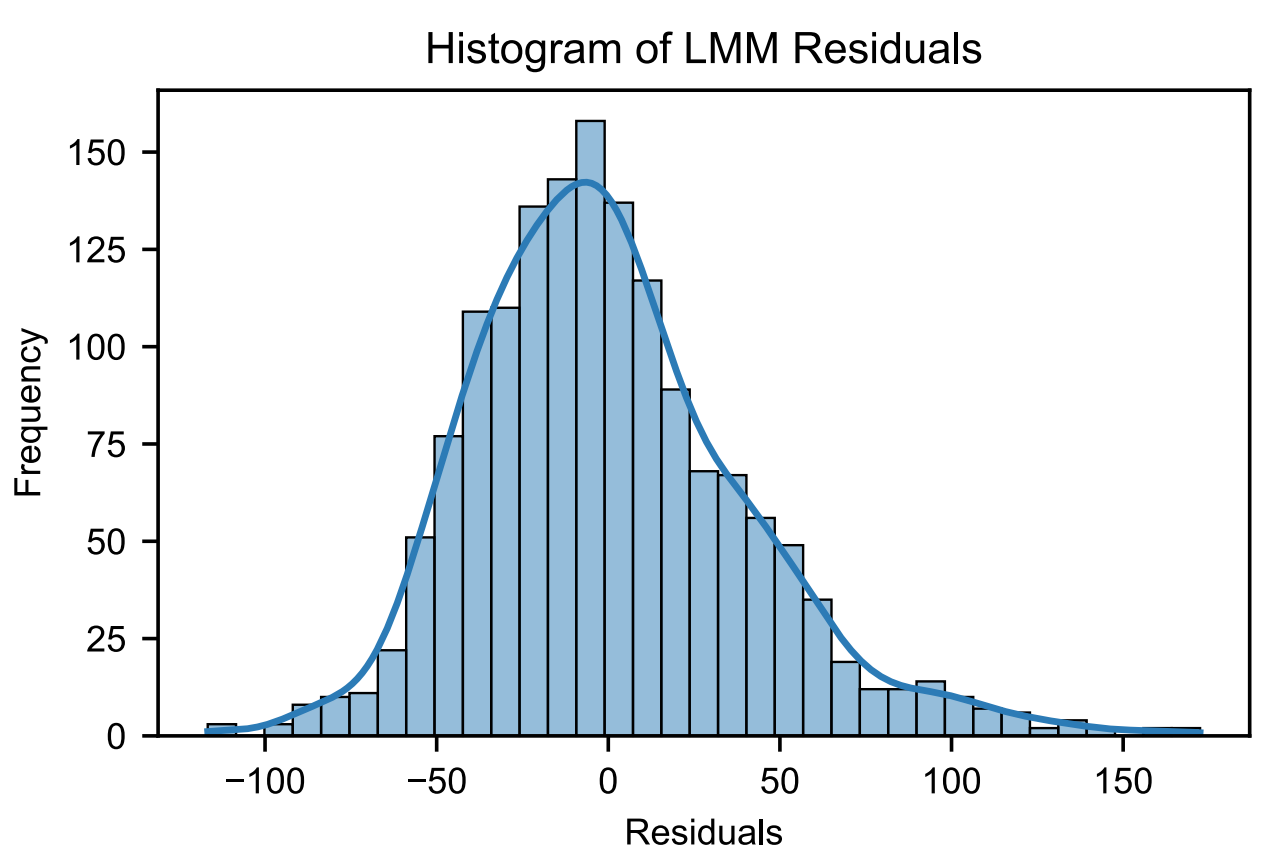

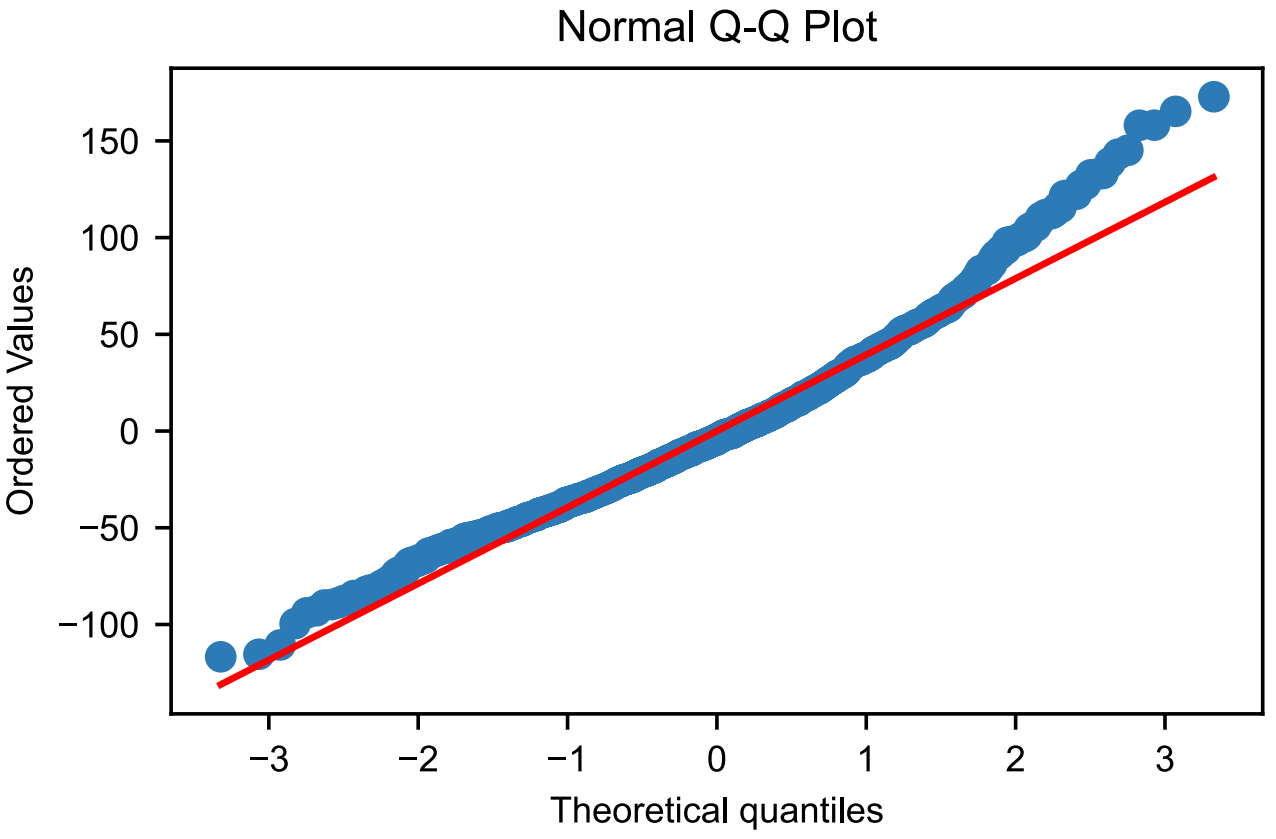

**b**

All Groups vs. Novice (In-context)

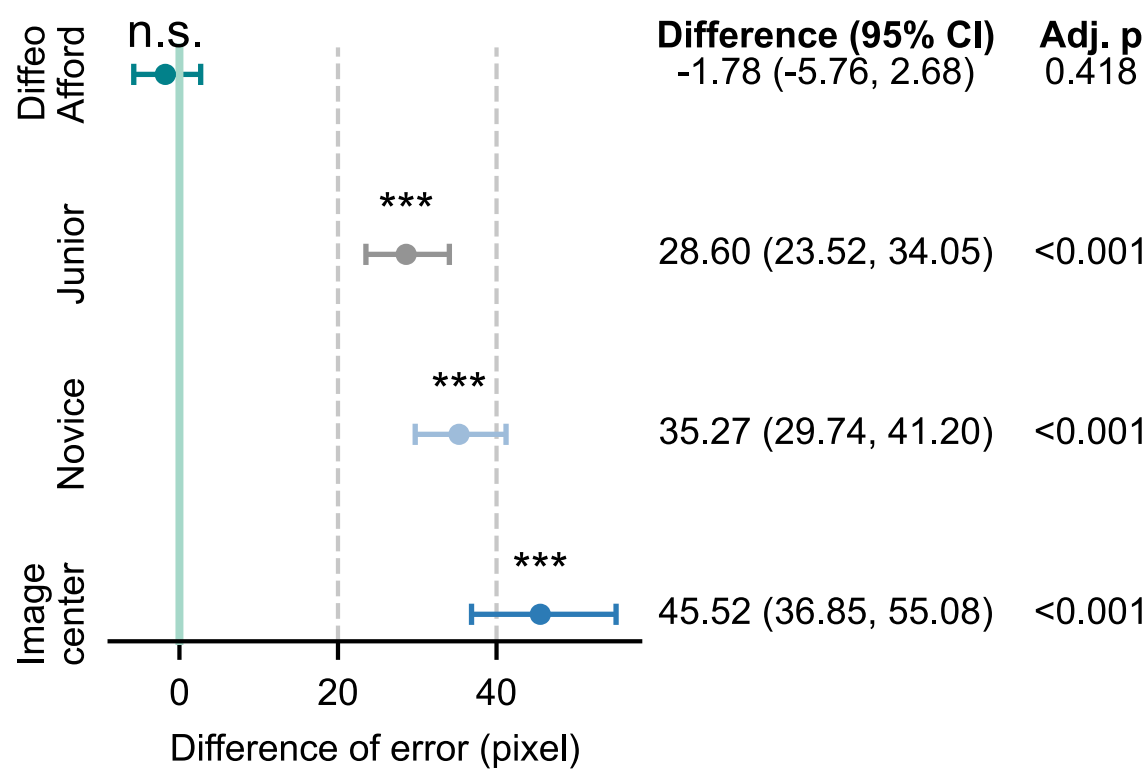

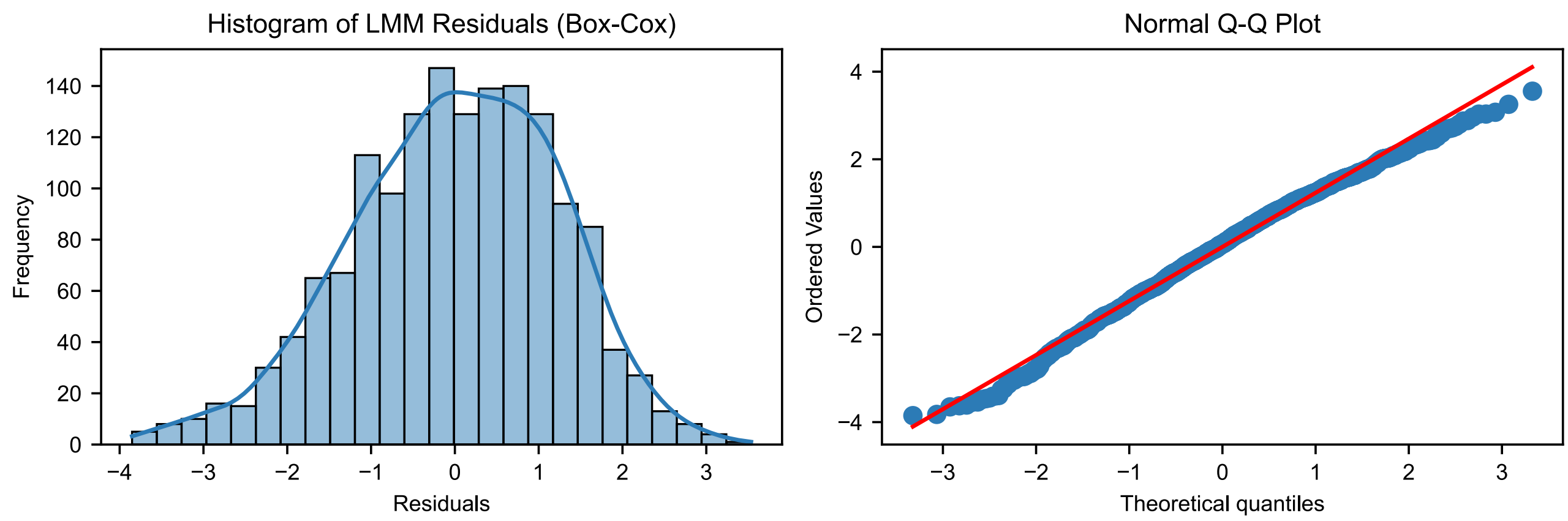

**c**

DiffeoAfford vs Novice (In-context)

Cholec80

Equivalent

−1.0 −0.5 0.0 0.5 1.0 1.5

Median error difference / Δ (90% bootstrap CI)

**Supplementary Fig. 2 | Inter-group annotation comparisons, residual diagnostics, and equivalence analysis on the Cholec80 dataset. a**, **b**, Statistical differences and model diagnostics evaluated before (**a**) and after (**b**) applying a Box-Cox transformation ($\lambda = 0.17$) to the annotation error data. In each panel, the top forest plot details the estimated differences in annotation errors relative to the Novice (In-context) reference group. The bottom panels display the corresponding residual diagnostics for the linear mixed models (LMMs), including the histogram of LMM residuals (left) and the normal Q-Q plot (right). The Box-Cox transformation applied in (**b**) effectively normalizes the right-skewed residual distribution observed in (**a**). **c**, Median-based bootstrap equivalence analysis comparing DiffeoAfford with Novice (In-context) across 141 paired images. The point and horizontal bar denote the normalized median paired error difference ($R$) and its 90% image-level bootstrap CI, respectively. The shaded region denotes the equivalence interval from −1 to 1; equivalence was concluded when the entire CI lay within this interval, which it did for DiffeoAfford versus Novice (In-context) ($R = 0.070$; 90% CI, −0.259 to 0.322). $R$ and the data-driven equivalence margin $\Delta$ are defined in Methods. The data-driven equivalence margin was $\Delta = 17.13$ pixels. LMM inference used two-sided asymptotic Wald z tests; finite denominator degrees of freedom are not defined. For the Cholec80 comparisons with Novice (In-context), $z = 13.649$ for the image-centre baseline, $z = 16.009$ for Novice, $z = 13.616$ for Junior and $z = -0.809$ for DiffeoAfford. CI, confidence interval; Adj. p, adjusted $p$ value; LMM, linear mixed model; Q-Q, quantile-quantile.

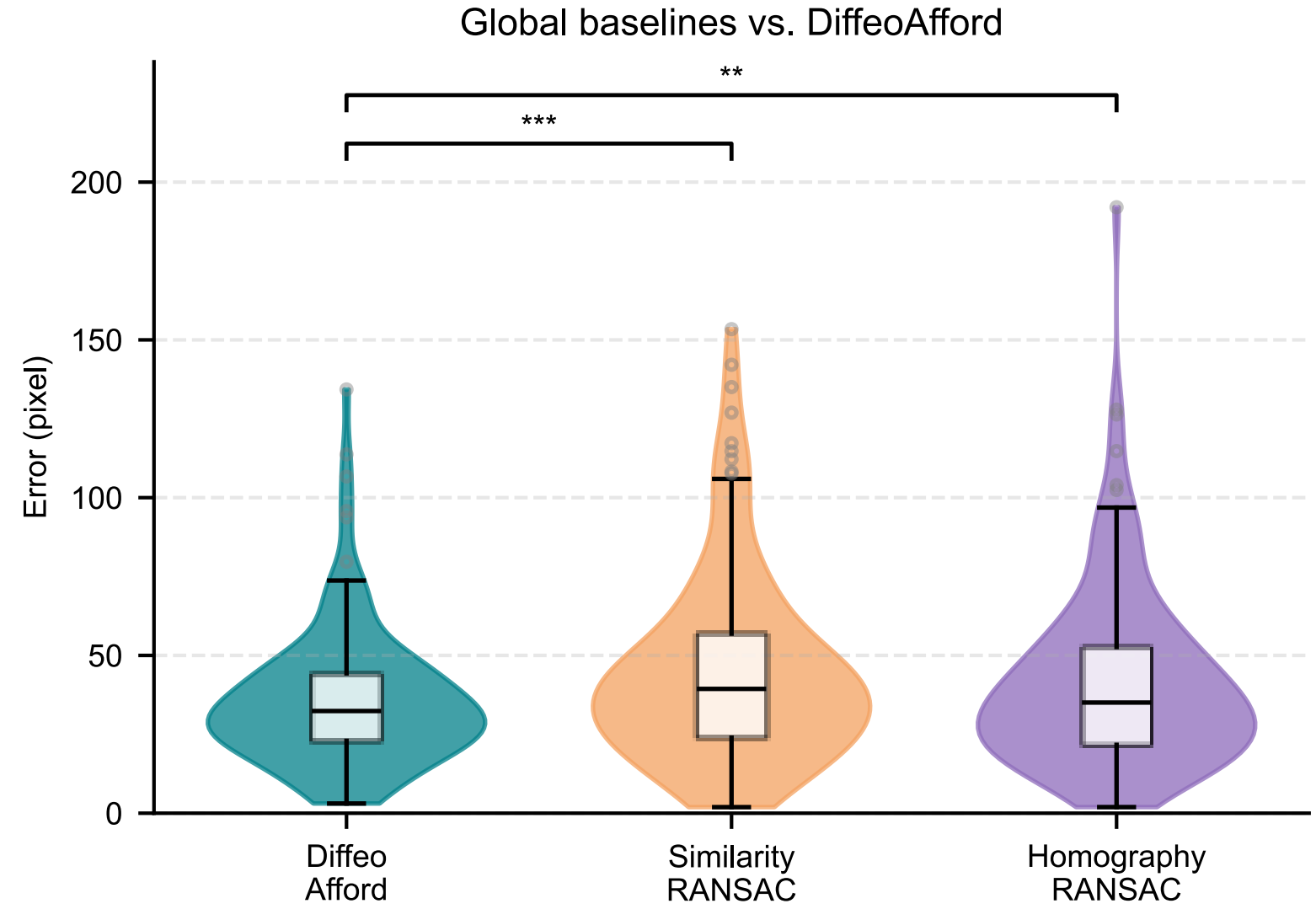

**Supplementary Fig. 3 | Comparison with global-transform affordance-grounding baselines.** Violin plots show grounding errors for DiffeoAfford, similarity RANSAC, and homography RANSAC[1, 3] across 141 paired Cholec80 images. Brackets indicate paired Wilcoxon signed-rank tests against DiffeoAfford after Benjamini–Hochberg correction. The Wilcoxon signed-rank statistics were W = 3056 for similarity RANSAC versus DiffeoAfford and $W = 3557$ for homography RANSAC versus DiffeoAfford ($n = 141$ paired images for each comparison); denominator degrees of freedom are not defined for these tests. **, $p < 0.01$; ***, $p < 0.001$.

**a**

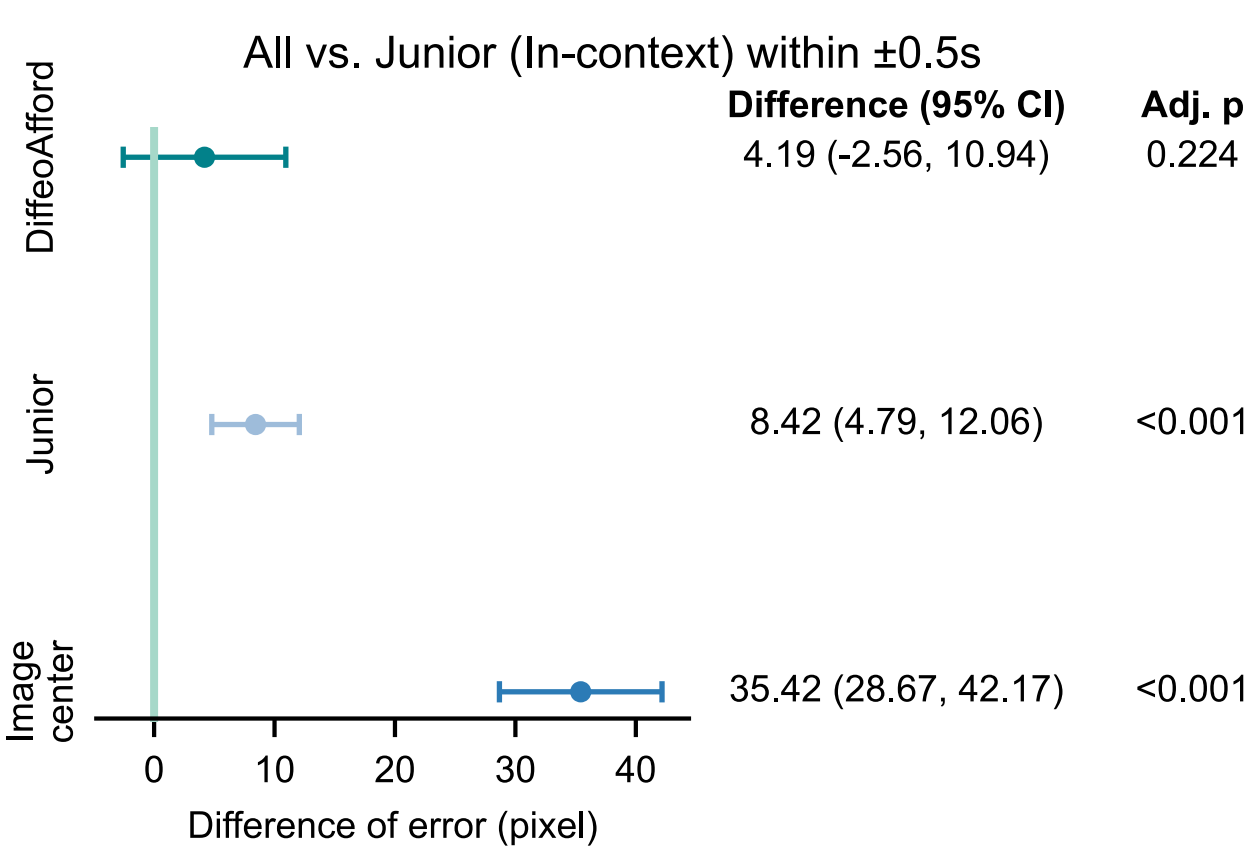

Residual Diagnostics (Time Window: ±0.5s)

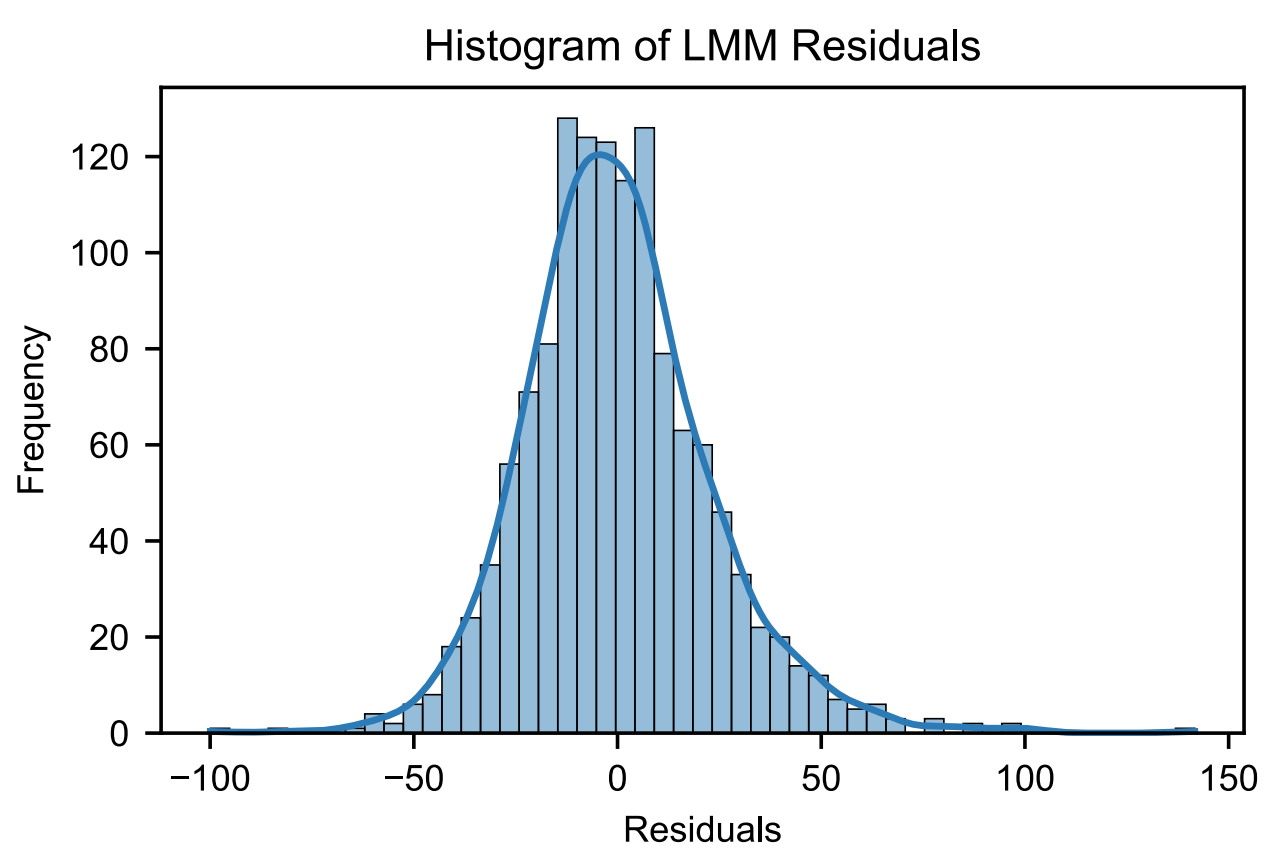

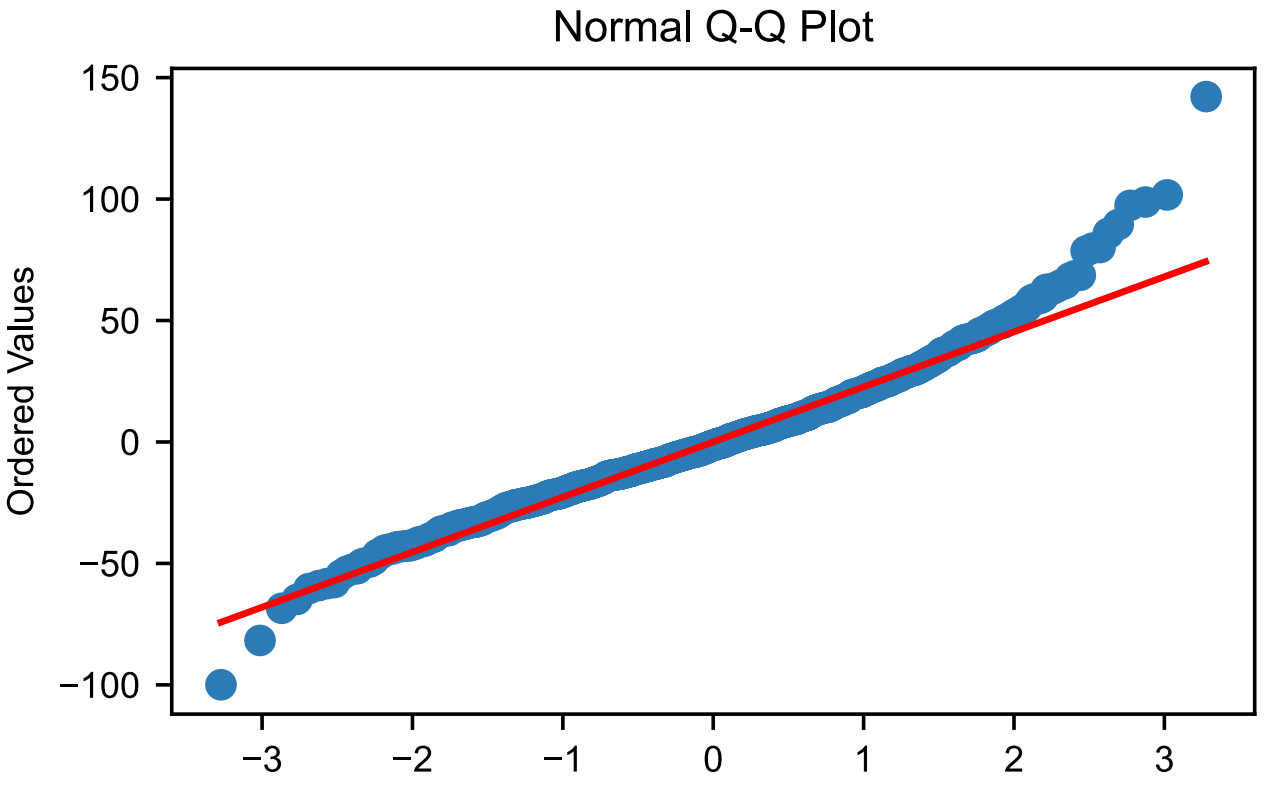

**b**

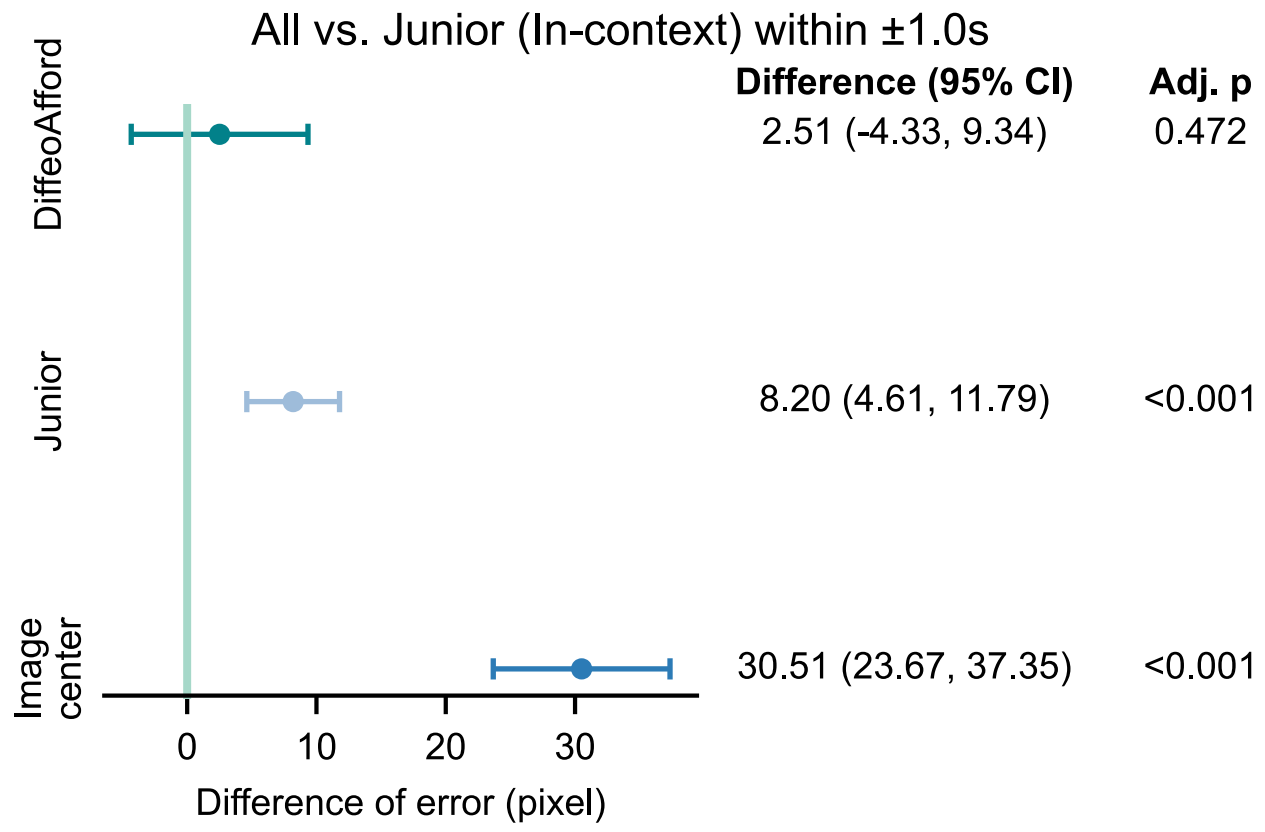

Residual Diagnostics (Time Window: ±1.0s)

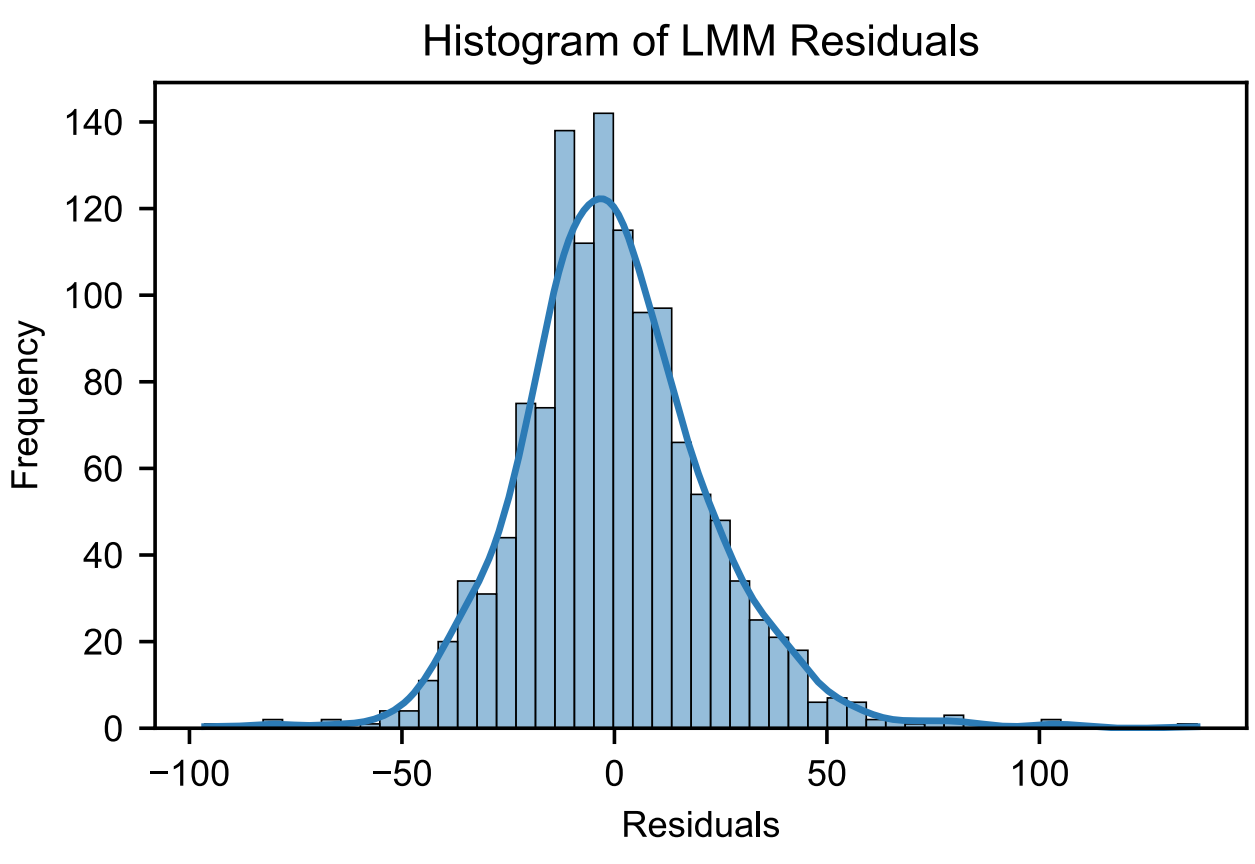

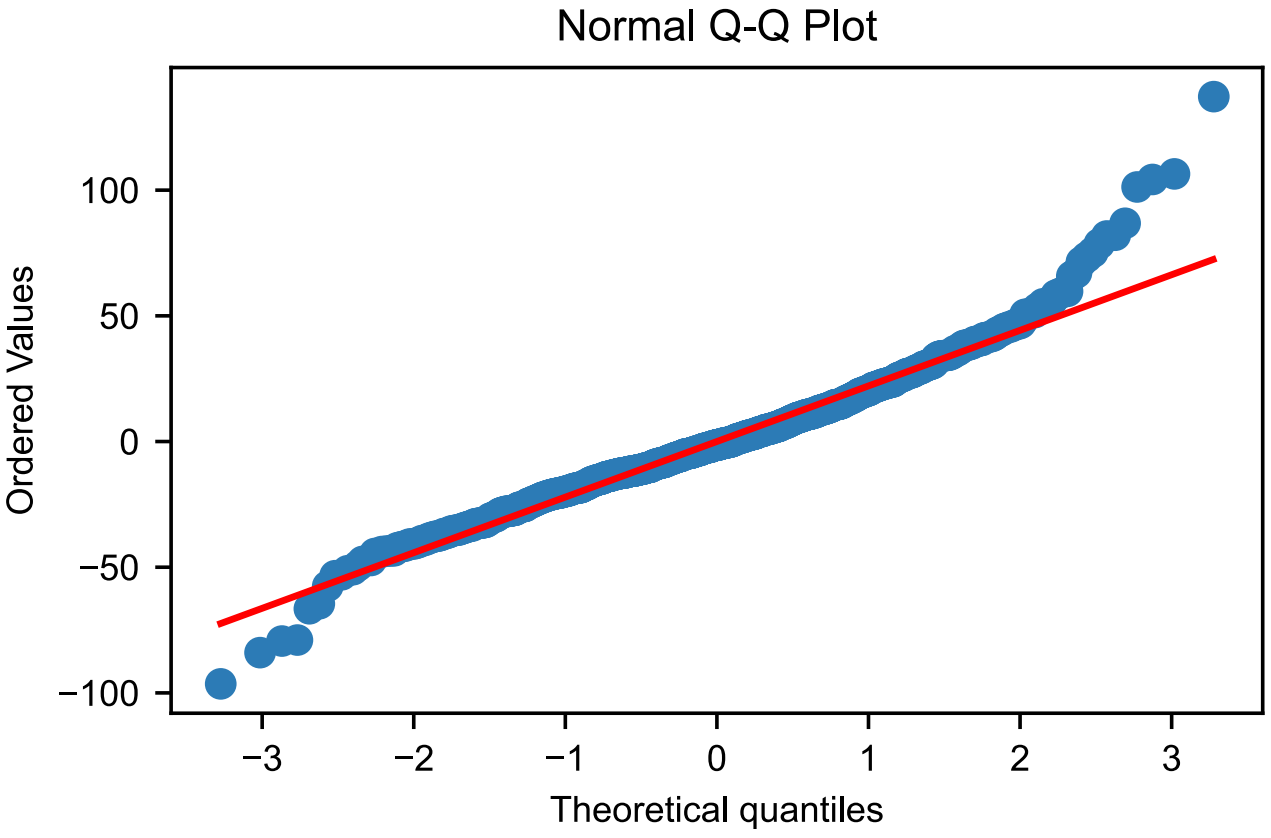

**c**

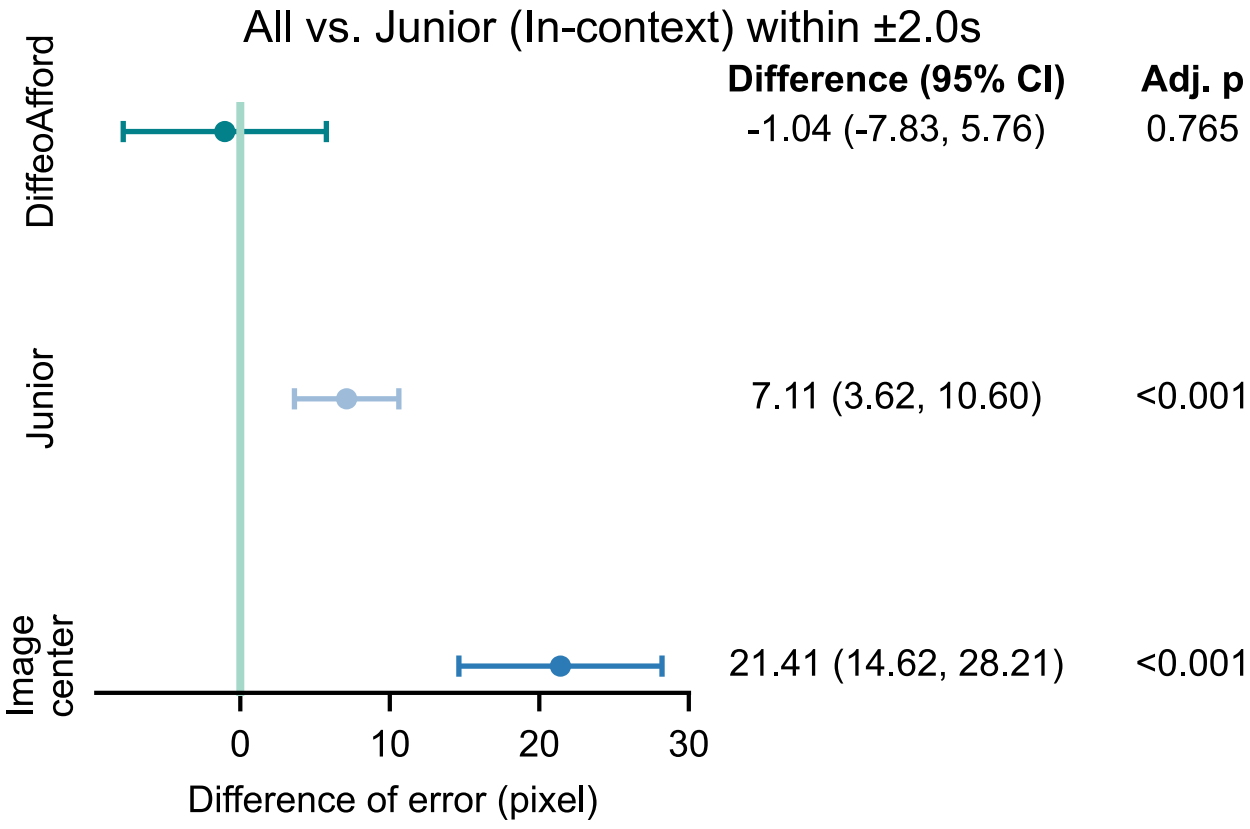

Residual Diagnostics (Time Window: ±2.0s)

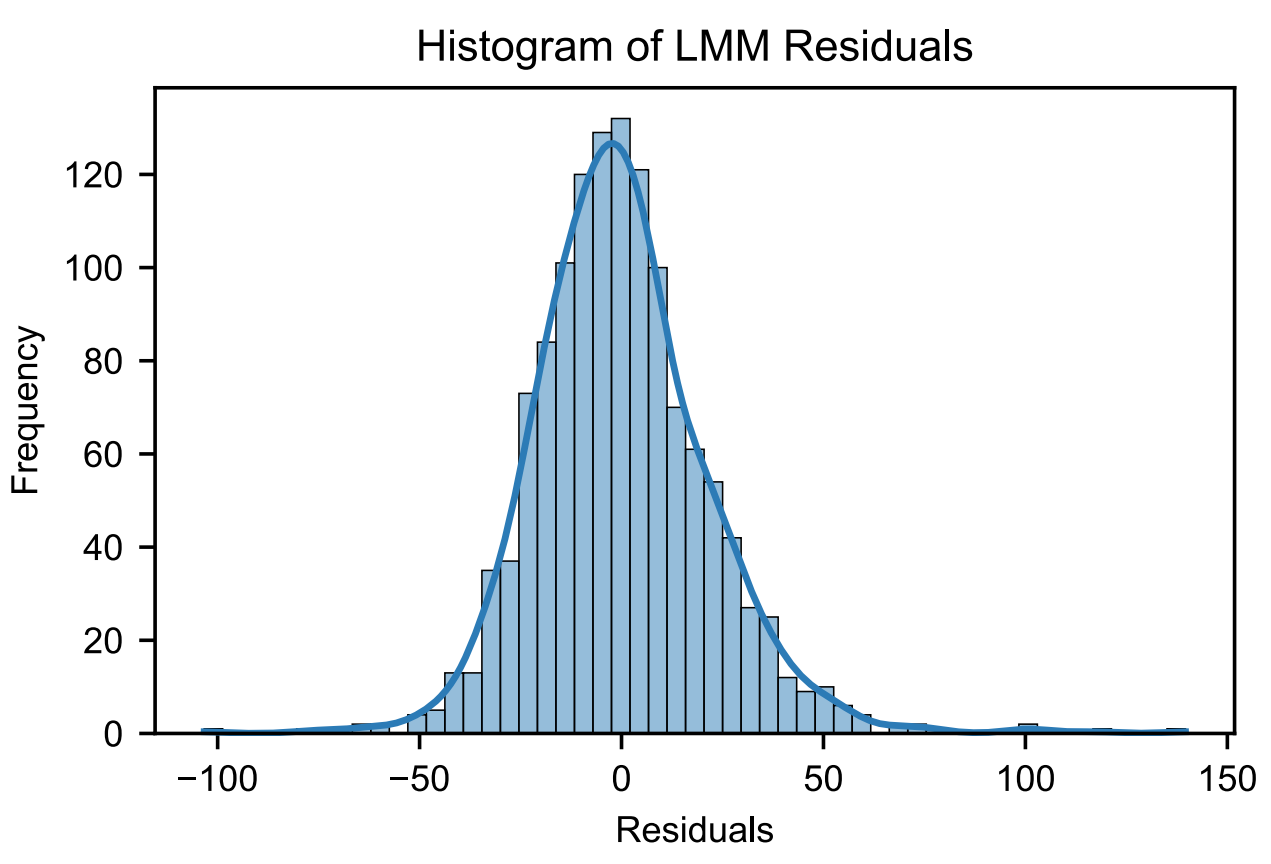

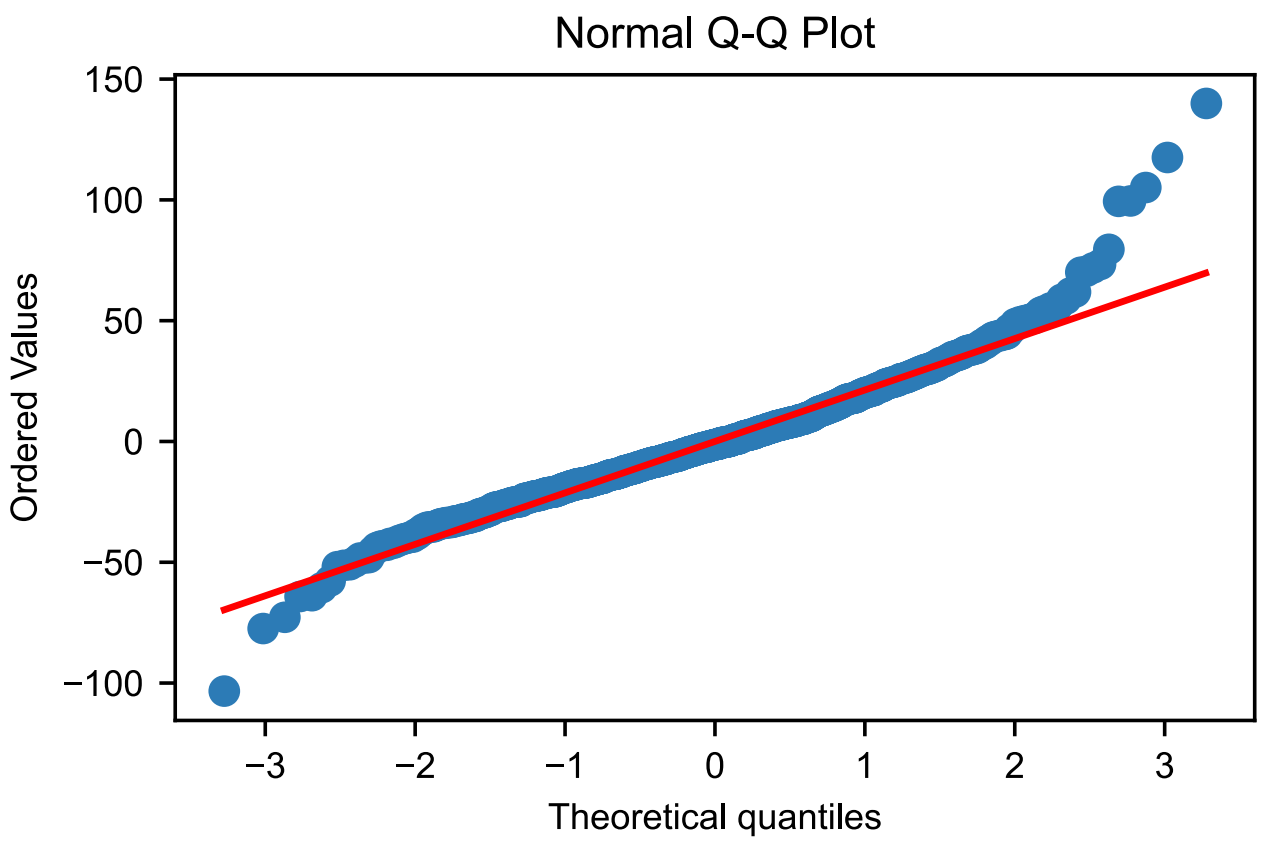

**d**

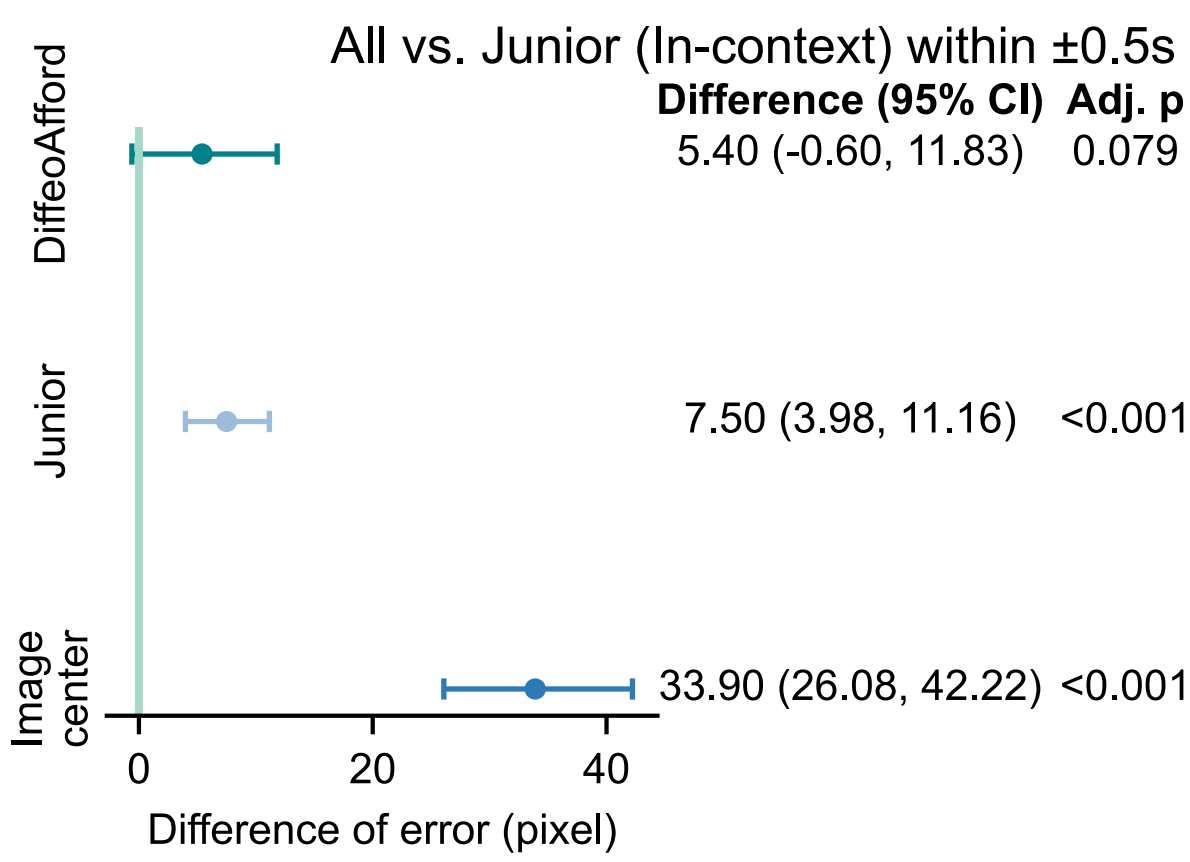

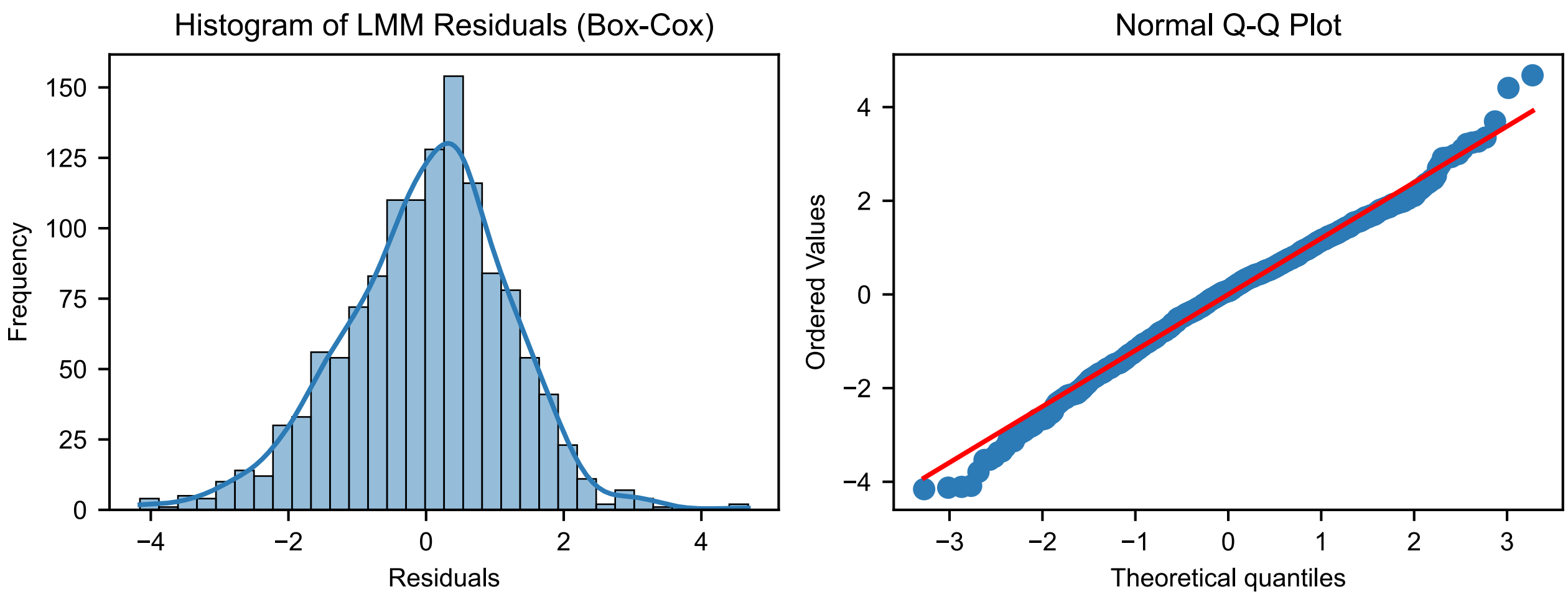

Residual Diagnostics (Time Window: ±0.5s, lambda=0.29)
Histogram of LMM Residuals (Box-Cox)
Frequency
Residuals
Normal Q-Q Plot
Ordered Values
Theoretical quantiles

**e**

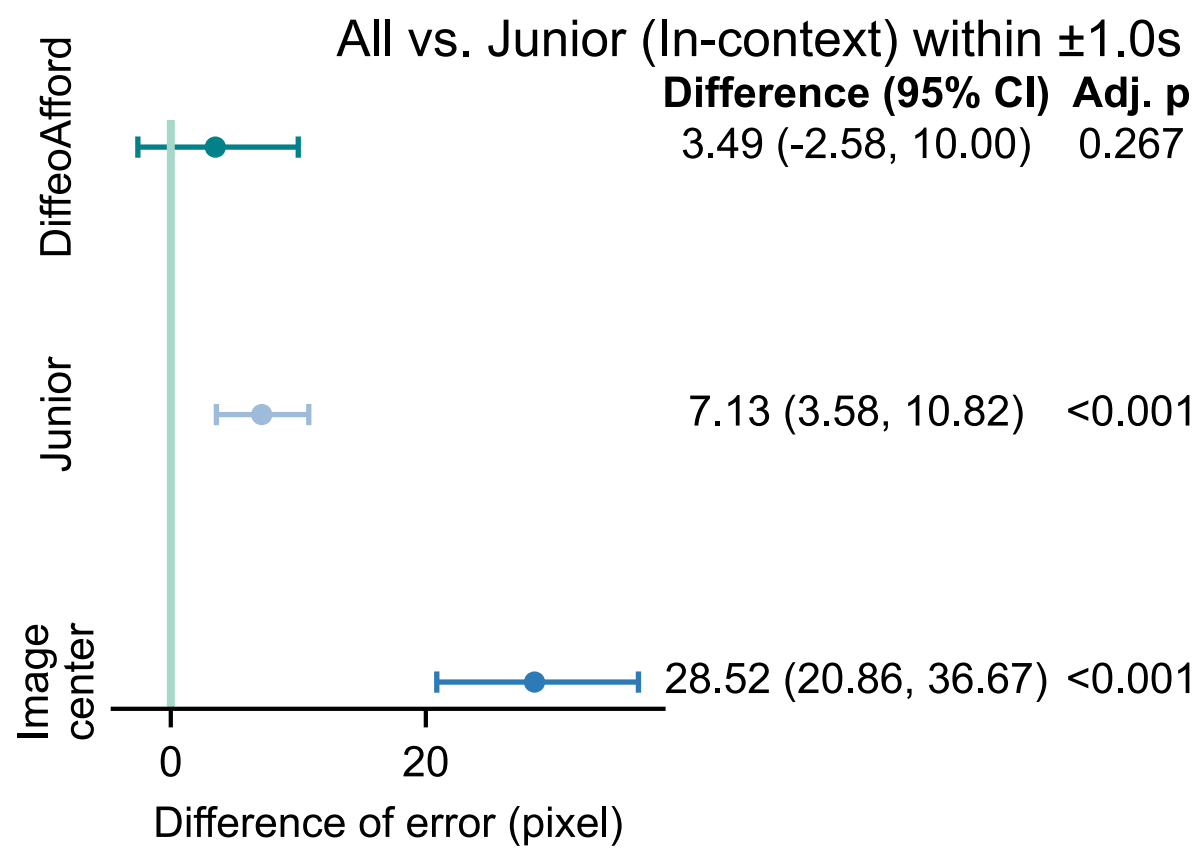

All vs. Junior (In-context) within ±1.0s
Difference (95% CI) Adj. p
DiffeoAfford
3.49 (-2.58, 10.00) 0.267
Junior
7.13 (3.58, 10.82) <0.001
Image center
28.52 (20.86, 36.67) <0.001
Difference of error (pixel)

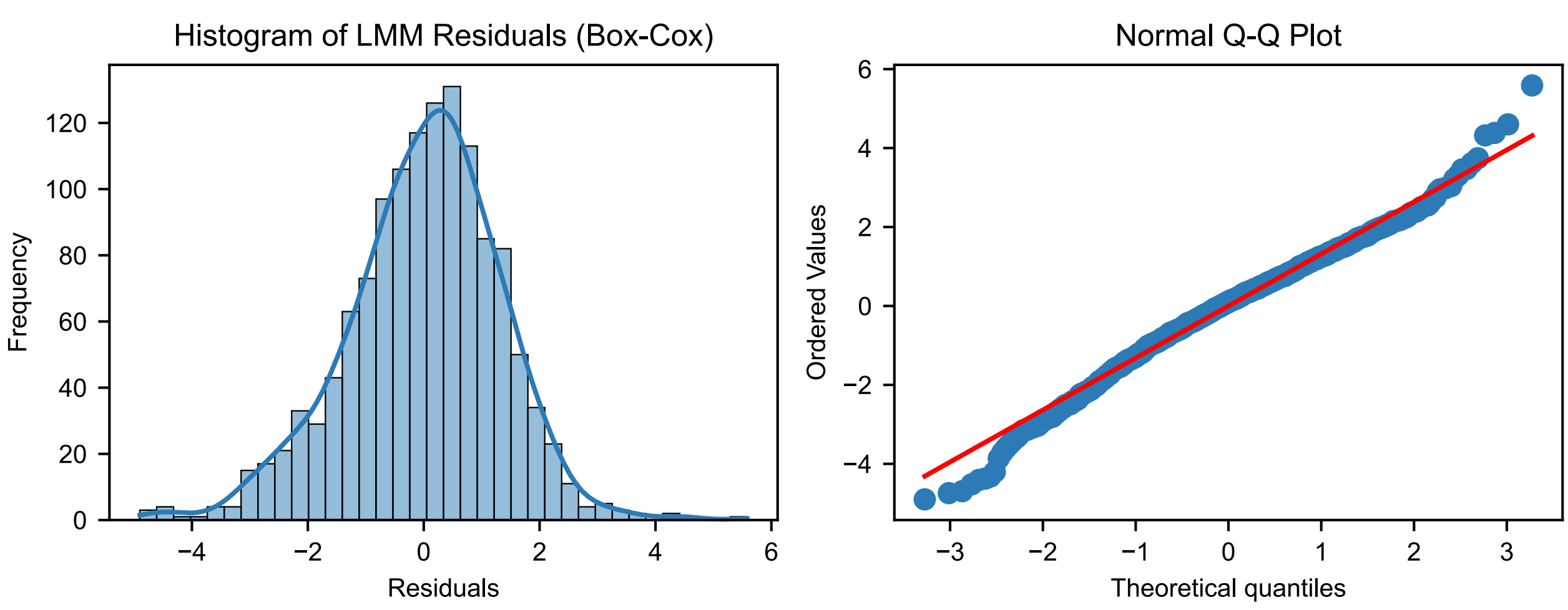

Residual Diagnostics (Time Window: ±1.0s, lambda=0.32)
Histogram of LMM Residuals (Box-Cox)
Frequency
Residuals
Normal Q-Q Plot
Ordered Values
Theoretical quantiles

**f**

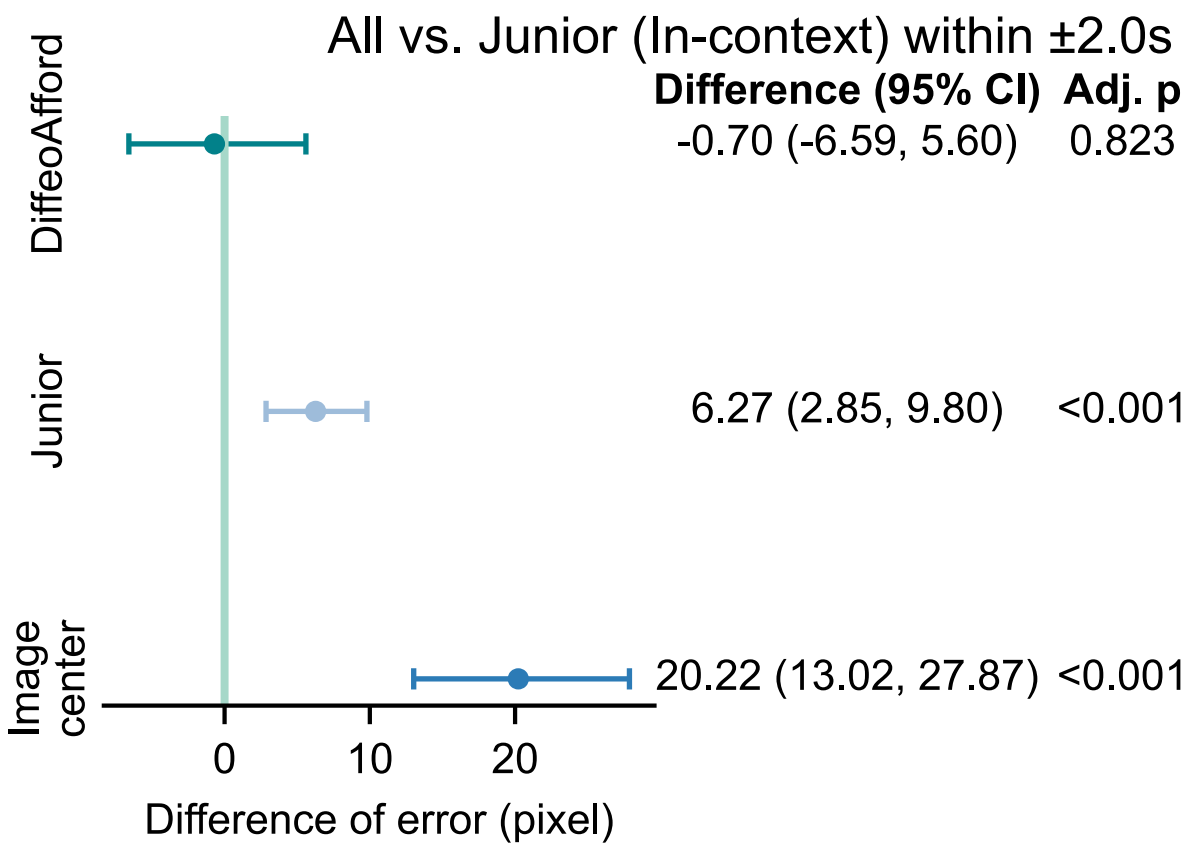

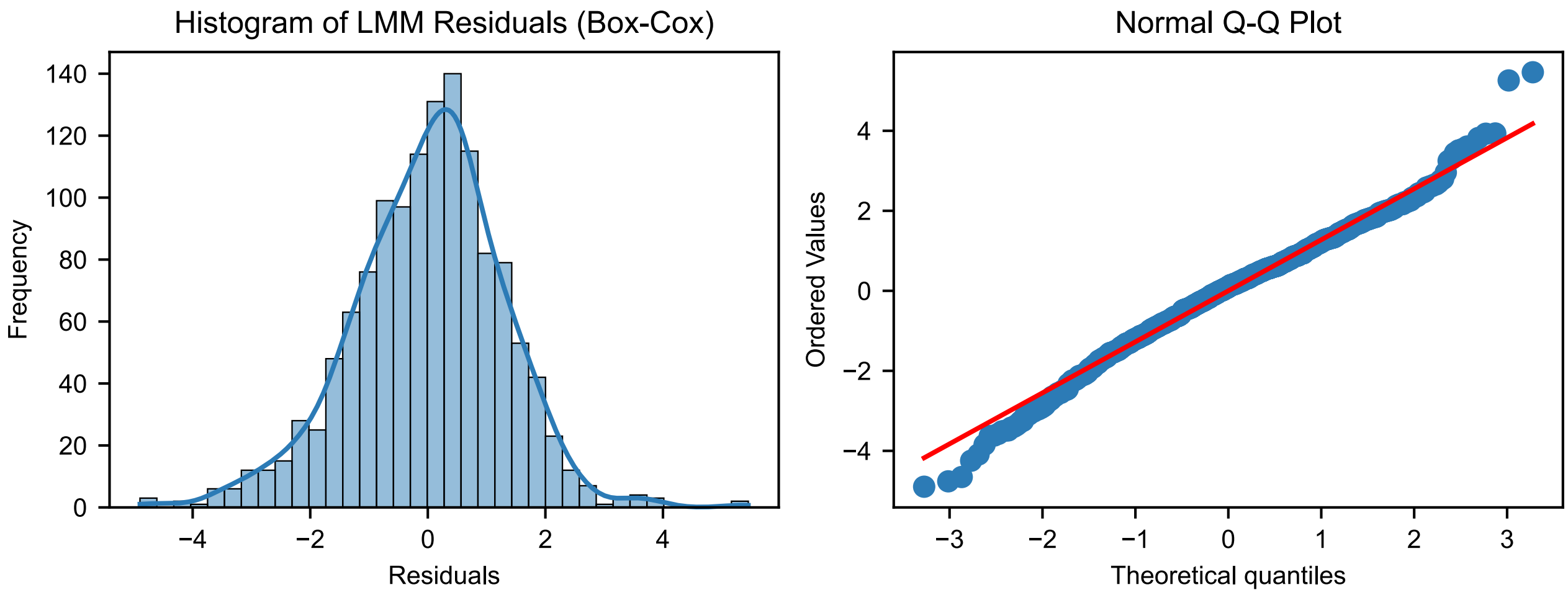

**g**

DiffeoAfford vs Junior (In-context)

LCET ±0.5 s — Equivalent

LCET ±1 s — Equivalent

LCET ±2 s — Equivalent

−1.0 −0.5 0.0 0.5 1.0 1.5

Median error difference / Δ (90% bootstrap CI)

**Supplementary Fig. 4 | Inter-group annotation comparisons, residual diagnostics, and equivalence analysis on the LCET dataset. a–f,** Statistical differences and model diagnostics evaluated across temporal windows of ±0.5 s (**a**, **d**), ±1.0 s (**b**, **e**), and ±2.0 s (**c**, **f**), presented before (**a–c**) and after (**d–f**) applying a Box-Cox transformation to the annotation error data. The optimal lambda parameters for the transformations are $\lambda = 0.29$ (**d**), $\lambda = 0.32$ (**e**), and $\lambda = 0.33$ (**f**). In each panel, the top forest plot details the estimated differences in annotation errors relative to the junior (in-context) reference group. The bottom panels display the corresponding residual diagnostics for the linear mixed models (LMMs), including the histogram of LMM residuals (left) and the normal Q-Q plot (right). The Box-Cox transformations applied in **d–f** effectively normalize the right-skewed residual distributions observed in **a–c**. **g**, Median-based bootstrap equivalence analyses

comparing DiffeoAfford with Junior (In-context) at the ±0.5-s, ±1.0-s, and ±2.0-s gaze windows. Points and horizontal bars denote $R$ and its 90% image-level bootstrap CI, respectively. The shaded region denotes the equivalence interval from −1 to 1; equivalence was concluded when the entire CI lay within this interval. The data-driven equivalence margins were $\Delta = 32.19, 28.98,$ and $29.62$ pixels for the ±0.5-s, ±1.0-s, and ±2.0-s windows, respectively. LMM inference used two-sided asymptotic Wald z tests; finite denominator degrees of freedom are not defined. Relative to Junior (In-context), the z statistics for the image-centre baseline, Junior and DiffeoAfford were 9.609, 4.269 and 1.757 at ±0.5 s; 8.041, 4.009 and 1.110 at ±1.0 s; and 5.853, 3.653 and −0.224 at ±2.0 s, respectively. CI, confidence interval; Adj. p, adjusted $p$ value; Q-Q, quantile-quantile.

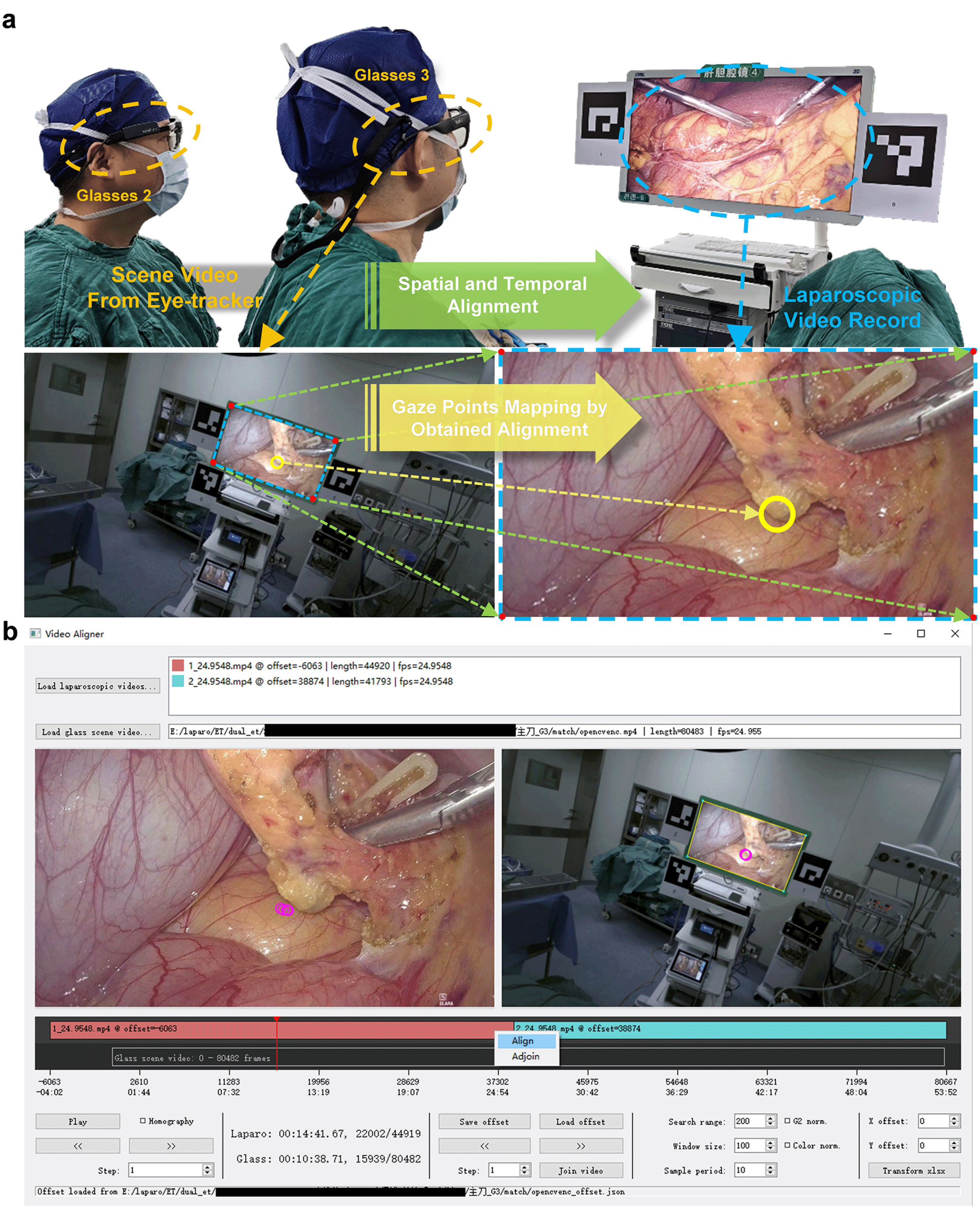

**Supplementary Fig. 5 | Framework for laparoscopic eye-tracking data collection. a**, Overview of the spatial

and temporal alignment process linking the scene video from the wearable eye-tracker with the laparoscopic video record, which enables mapping of the surgeon's gaze points onto the laparoscopic video. **b**, The software developed for performing the spatial and temporal alignment of the videos.

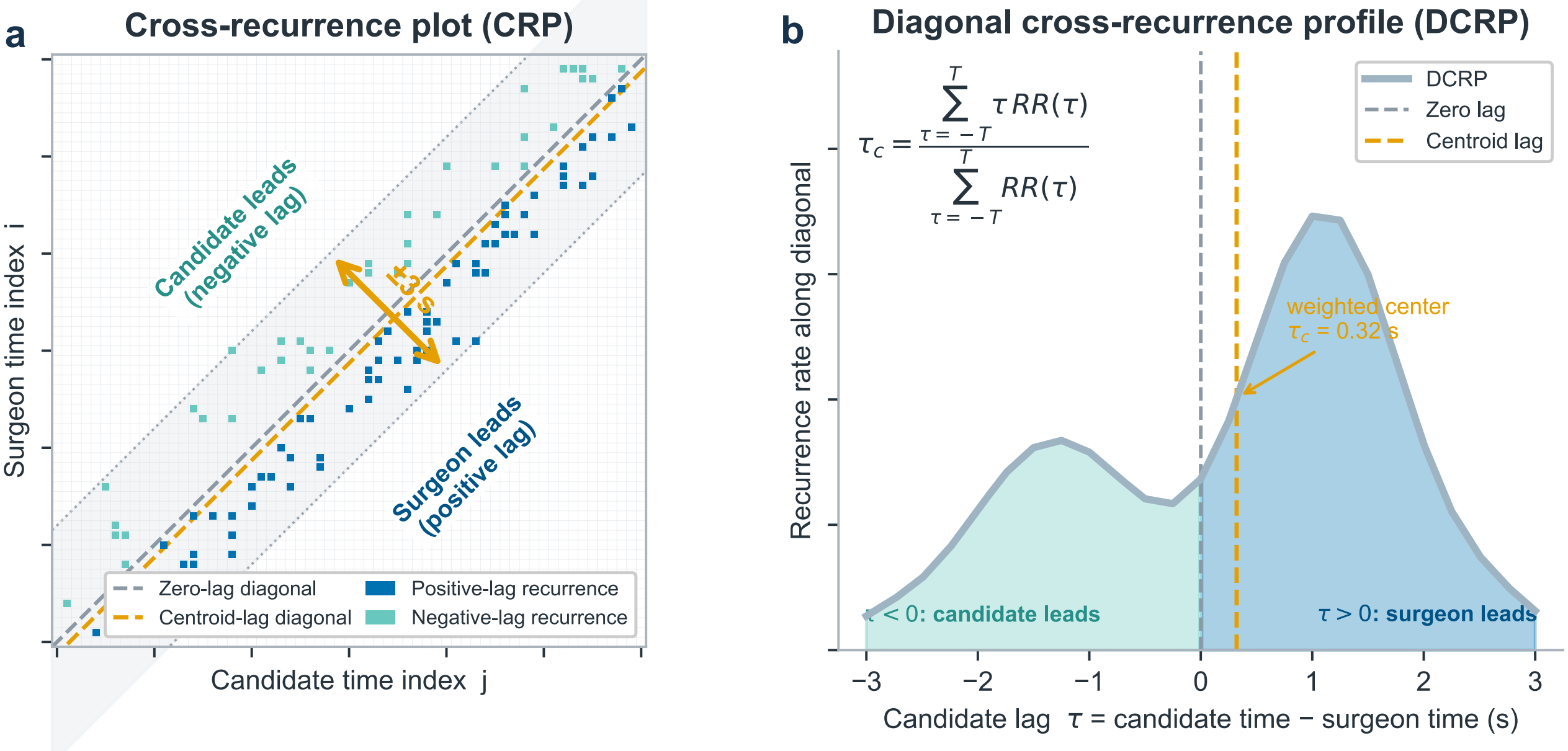

$$\tau_c = \frac{\sum_{\tau=-T}^{T} \tau RR(\tau)}{\sum_{\tau=-T}^{T} RR(\tau)}$$

**Supplementary Fig. 6 | Principle of cross-recurrence quantification analysis (CRQA), the diagonal cross-recurrence profile (DCRP), and centroid lag. a**, Cross-recurrence plot comparing surgeon gaze at time index $i$ with a candidate sequence—either AH prediction or camera-assistant gaze—at time index $j$. A recurrence is recorded when the angular separation between the two sequences does not exceed $1.5°$. Only recurrences with candidate lag $-3\,s \le \tau \le +3\,s$ are included. Cells above the main diagonal represent negative candidate lags, for which the candidate leads surgeon gaze; cells below it represent positive candidate lags, for which the surgeon leads. **b**, Recurrences sharing the same candidate lag $\tau$ are aggregated to form the DCRP $RR(\tau)$.[6] The centroid lag $\tau_c$ is the recurrence-rate-weighted mean of $\tau$ across the prespecified lag range.[7] Negative values indicate that the candidate tends to lead surgeon gaze, positive values indicate that the surgeon tends to lead the candidate, and values near zero indicate temporal coupling without a consistent lead or lag. Recurrence cells and profiles are schematic and do not represent study data.

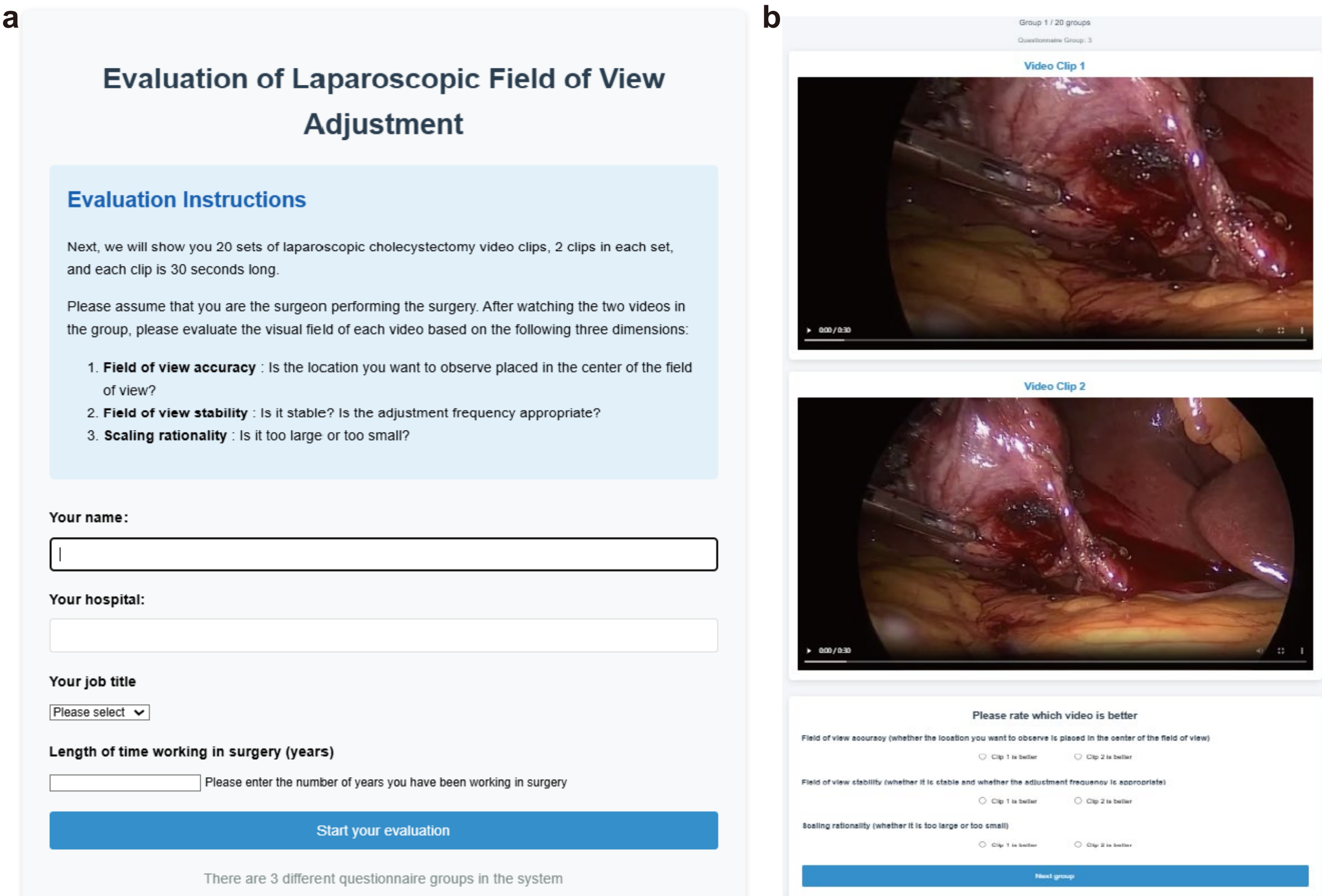

**Supplementary Fig. 7 | User interface of the online multicenter questionnaire survey. a**, The registration and instruction page outlining the evaluation criteria. **b**, The paired video evaluation page, where participants simultaneously viewed two synchronized video clips (adjusted via different methods) and indicated their preference across three specified dimensions.

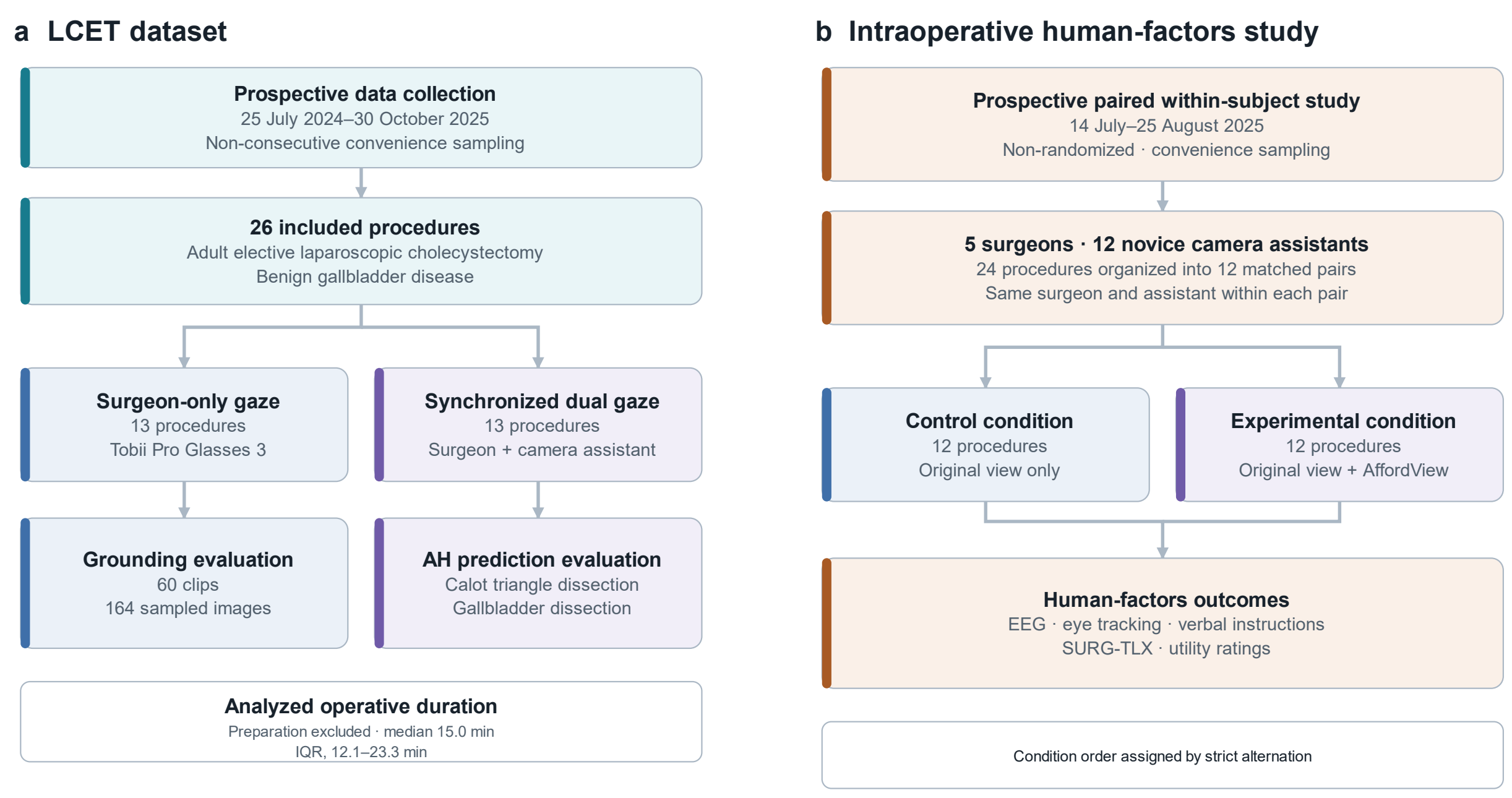

**Supplementary Fig. 8 | Prospective data collection and study flow. a**, Flow of the LCET dataset, showing prospective non-consecutive convenience sampling, included procedures, gaze-recording subgroups, and dataset-specific evaluations. **b**, Flow of the prospective paired within-subject human-factors study, showing matched-pair allocation, control and experimental conditions, and assessed outcomes. AH, affordance hotspot; EEG, electroencephalography; IQR, interquartile range; LCET, laparoscopic cholecystectomy eye tracking; SURG-TLX, Surgery Task Load Index.

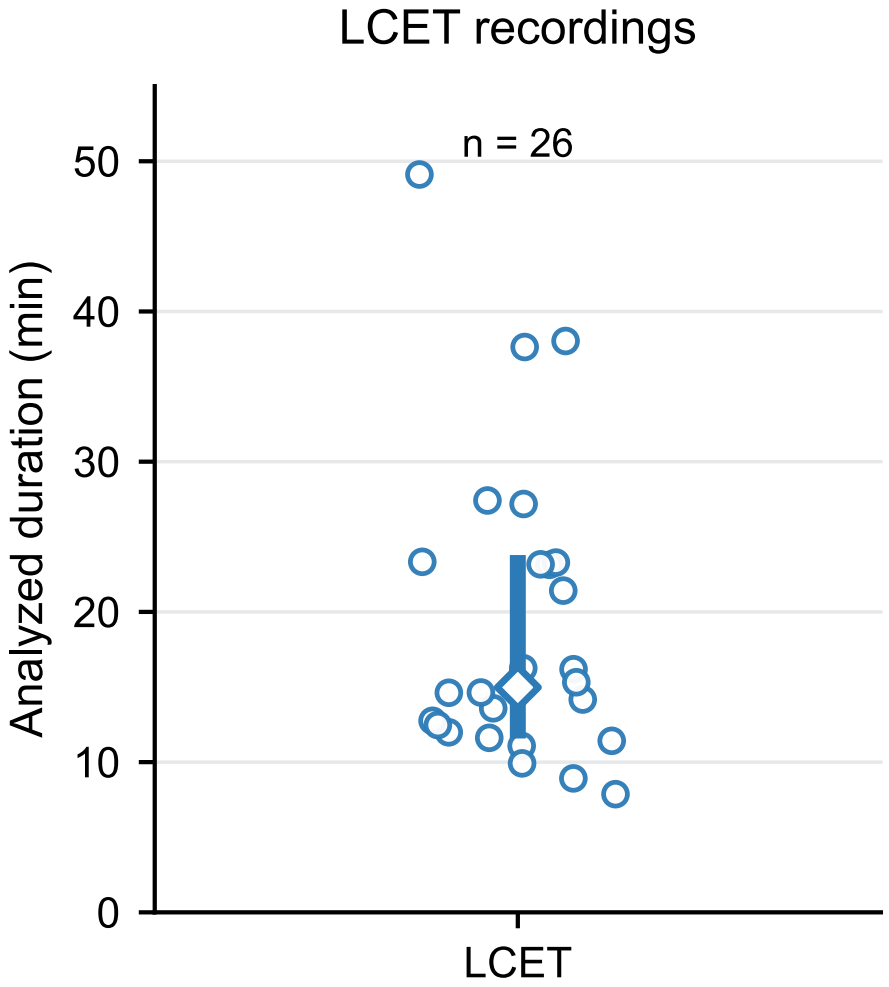

**Supplementary Fig. 9 | Analyzed operative duration in the LCET dataset.** Each point represents one of 26 procedures. Duration was calculated by summing the annotated phase intervals after excluding Preparation; unannotated gaps were not counted. The diamond and vertical bar denote the median and interquartile range, respectively (median, 15.0 min; IQR, 12.1–23.3 min).

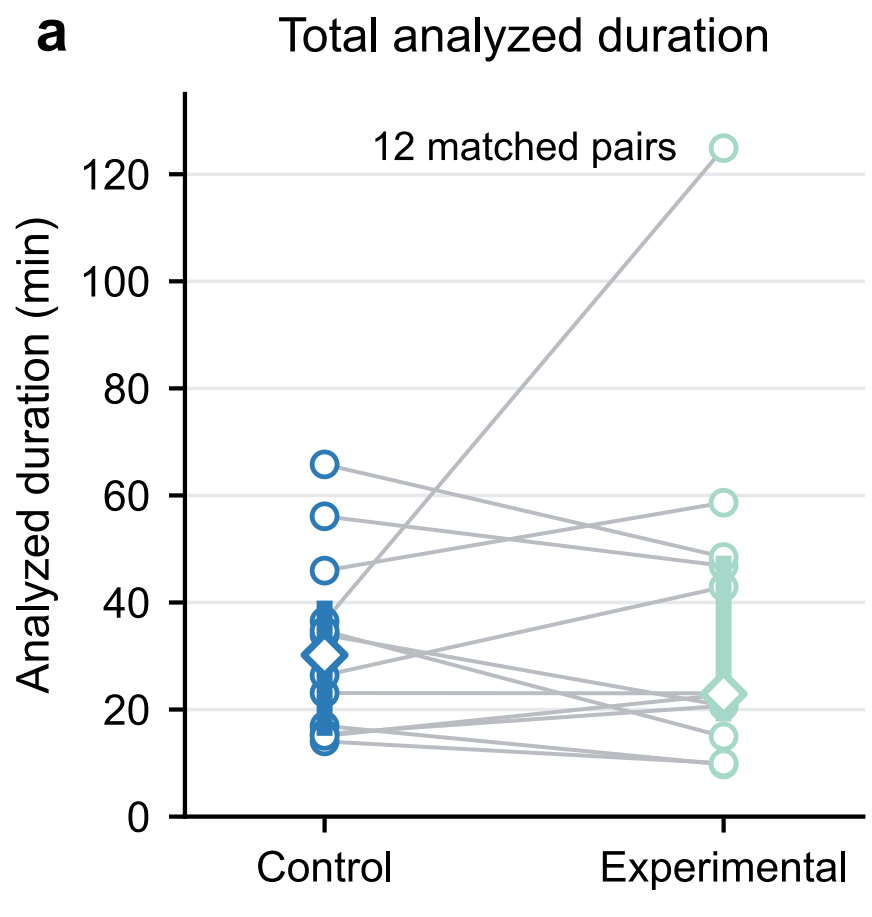

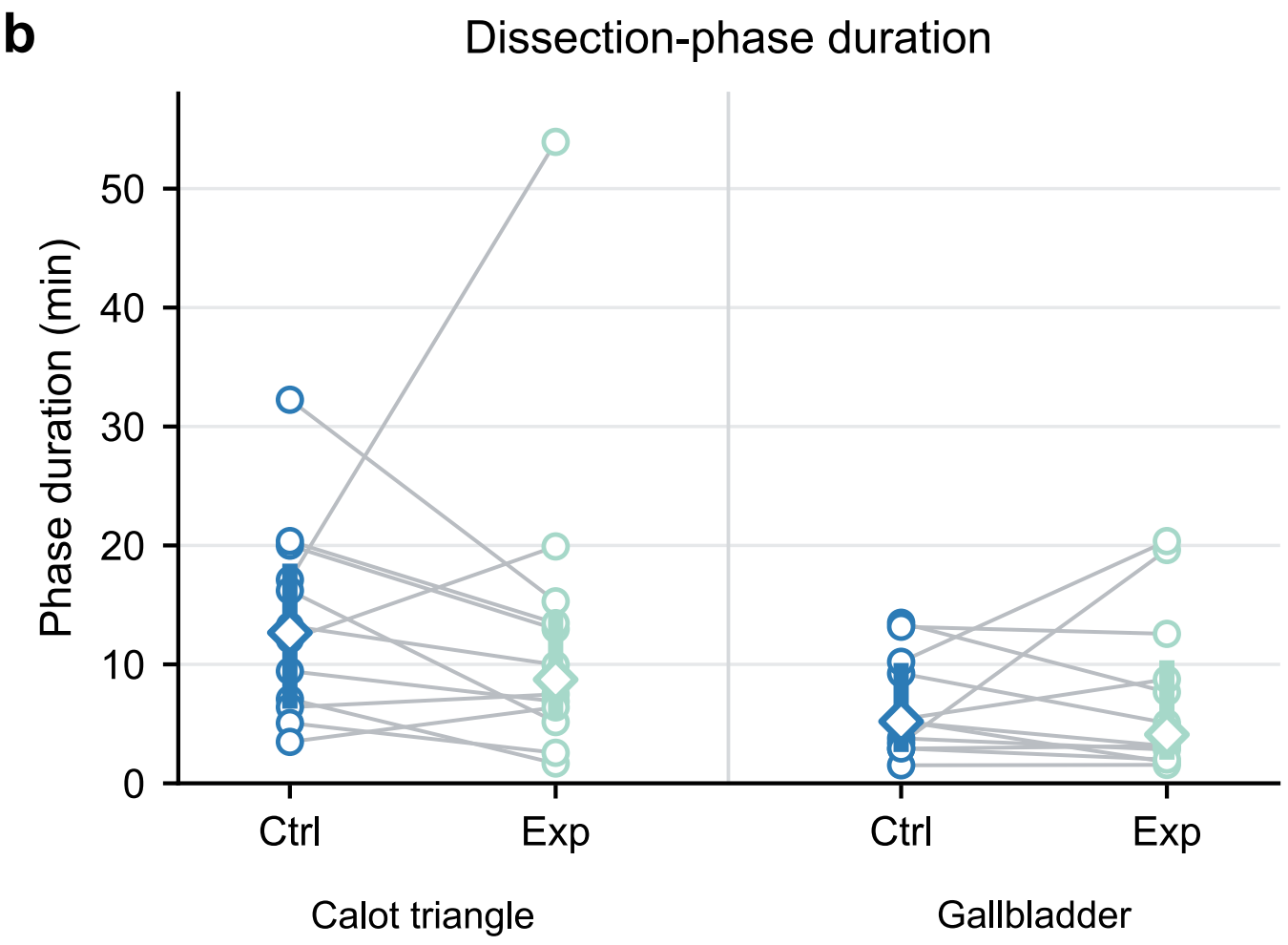

**Supplementary Fig. 10 | Analyzed operative duration in the intraoperative human-factors study. a**, Paired total analyzed duration in the control and experimental conditions across 12 matched pairs. Total duration was calculated by summing all annotated phase intervals, including Preparation; unannotated gaps were not counted. **b**, Paired durations of Calot triangle dissection and gallbladder dissection. Gray lines connect procedures within matched pairs; diamonds and vertical bars denote medians and interquartile ranges, respectively. Ctrl, control; Exp, experimental.